\documentclass[final]{msml2020} % Include author names

\title[Geometric Scattering Networks on Manifolds]{Geometric Wavelet Scattering Networks \\ on Compact Riemannian Manifolds}
\usepackage{times}

\msmlauthor{%
 \Name{Michael Perlmutter} \Email{perlmut6@msu.edu} \\
 \addr  Michigan State University \\ 
        Department of Computational Mathematics, Science \& Engineering \\ 
        East Lansing, Michigan, USA
 \AND
 \Name{Feng Gao} \Email{f.gao@yale.edu} \\
 \addr  Yale University \\
        Department of Genetics \\
        New Haven, Connecticut, USA
 \AND
 \Name{Guy Wolf} \Email{guy.wolf@umontreal.ca} \\
 \addr  Universit\'{e} de Montr\'{e}al \\
        Department of Mathematics and Statistics \\
        Mila -- Quebec Artificial Intelligence Institute \\
        Montr\'{e}al, Qu\'{e}bec, Canada
 \AND
 \Name{Matthew Hirn} \Email{mhirn@msu.edu} \\
 \addr  Michigan State University \\
        Department of Computational Mathematics, Science \& Engineering \\
        Department of Mathematics \\
        Center for Quantum Computing, Science \& Engineering \\
        East Lansing, Michigan, USA
}

\usepackage{color}
\usepackage{bm}
\usepackage{thmtools}
\usepackage{comment}
\usepackage{mathtools}
\usepackage{adjustbox}
\usepackage{tikz}
\usetikzlibrary{positioning}
\usepackage{wrapfig}
\usepackage{tabu}
\usepackage{hhline}

\declaretheorem[name=Theorem, numberwithin=section]{thm}
\declaretheorem[name=Proposition, sibling=thm]{prop}
\declaretheorem[name=Lemma, sibling=thm]{lem}

\newcommand{\Diff}{\mathrm{Diff}}
\newcommand{\ellb}{\bm{\ell}}
\newcommand{\hf}{\widehat{f}}
\newcommand{\hh}{\widehat{h}}
\newcommand{\hphi}{\widehat{\phi}}
\newcommand{\hpsi}{\widehat{\psi}}
\newcommand{\id}{\mathrm{id}}
\newcommand{\Isom}{\mathrm{Isom}}
\newcommand{\vol}{\mathrm{vol}}
\newcommand{\Lb}{\mathbf{L}}
\newcommand{\Mc}{\mathcal{M}}
\newcommand{\N}{\mathbb{N}}
\newcommand{\R}{\mathbb{R}}
\newcommand{\Wc}{W}

\newcommand{\Z}{\mathbb{Z}}
\newcommand{\Cb}{\mathbf{C}}

\newcommand{\Pc}{\mathcal{P}}
\newcommand{\Sbar}{\overline{S}}
\newcommand{\diam}{\mathrm{diam}}

\begin{document}

\maketitle

\begin{abstract}
  The Euclidean scattering transform was introduced nearly a decade ago to improve the mathematical understanding of convolutional neural networks. Inspired by recent interest in geometric deep learning, which aims to generalize convolutional neural networks to manifold and graph-structured domains, we define a geometric scattering transform on manifolds. Similar to the Euclidean scattering transform, the geometric scattering transform is based on a cascade of wavelet filters and pointwise nonlinearities. It is invariant to local isometries and stable to certain types of diffeomorphisms. Empirical results demonstrate its utility on several geometric learning tasks. Our results generalize the deformation stability and local translation invariance of Euclidean scattering, and demonstrate the importance of linking the used filter structures to the underlying geometry of the data. 
\end{abstract}


\section{Introduction}

In an effort to improve our mathematical understanding of deep convolutional networks and their learned features, \citet{mallat:firstScat2010,mallat:scattering2012} introduced the \emph{scattering transform} for signals on $\R^d$. This transform has an architecture similar to convolutional neural networks (ConvNets), based on a cascade of convolutional filters and simple pointwise nonlinearities. However, unlike many other deep learning methods, this transform uses the complex modulus as its nonlinearity and does not learn its filters from data, but instead uses designed filters. As shown in~\cite{mallat:scattering2012}, with properly chosen wavelet filters, the scattering transform is provably invariant to the actions of certain Lie groups, such as the translation group, and is also provably Lipschitz stable to small diffeomorphisms, where the size of a diffeomorphism is quantified by its deviation from a translation. These notions were applied in \cite{bruna:scatClass2011, bruna:invariantScatConvNet2013, sifre:rotoScatTexture2012, mallat:rotoScat2013, mallat:rigidMotionScat2014, oyallon:scatObjectClass2014} using groups of translations, rotations, and scaling operations, with applications in image and texture classification. Additionally, the scattering transform and its deep filter bank approach have also proven to be effective in several  other fields, such as audio processing \citep{anden:scatAudioClass2011, anden:deepScatSpectrum2014, wolf:BSS-mlsp, wolf:BSS, Anden2019JointTimeFrequency}, medical signal processing \citep{talmon:scatManifoldHeart2014}, and quantum chemistry~\citep{hirn:waveletScatQuantum2016, eickenberg:3DSolidHarmonicScat2017, eickenberg:scatMoleculesJCP2018, brumwell:steerableScatLiSi2018}. Mathematical generalizations to non-wavelet filters have also been studied, including Gabor filters as in the short time Fourier transform \citep{czaja:timeFreqScat2017} and more general classes of semi-discrete frames \citep{grohs:cnnCartoonFcns2016, wiatowski:frameScat2015, wiatowski:mathTheoryCNN2018}.

However, many data sets of interest have an intrinsically non-Euclidean structure and are  better modeled by graphs or manifolds. Indeed, manifold learning models \cite[e.g.,][]{tenenbaum:isomap2000,coifman:diffusionMaps2006,maaten:tSNE2008} are commonly used for representing high-dimensional data in which unsupervised algorithms infer data-driven geometries to capture intrinsic structure in data.
Furthermore, signals supported on manifolds are becoming increasingly prevalent, for example, in shape matching and computer graphics. As such, a large body of work has emerged to explore the generalization of spectral and signal processing notions to manifolds \citep{coifman:geometricHarmonics2006} and graphs \cite[and references therein]{shuman:emerging2013}. In these settings, functions are supported on the manifold or the vertices of the graph, and the eigenfunctions of the Laplace-Beltrami operator, or the eigenvectors of the graph Laplacian, serve as the Fourier harmonics. This increasing interest in non-Euclidean data geometries has led to a new research direction known as \emph{geometric deep learning}, which aims to generalize convolutional networks to graph and manifold structured data~\cite[and references therein]{Bronstein:geoDeepLearn2017}. 

Inspired by geometric deep learning, recent works have also proposed an extension of the scattering transform to graph domains. These mostly focused on finding features that represent a graph structure (given a fixed set of signals on it) while being stable to graph perturbations. In \cite{gama:diffScatGraphs2018}, a cascade of diffusion wavelets from~\cite{coifman:diffWavelets2006} was proposed, and its Lipschitz stability was shown with respect to a global diffusion-inspired distance between graphs. These results were generalized in \cite{gama:stabilityGraphScat2019} to graph wavelets constructed from more general graph shift operators.  A similar construction discussed in \cite{zou2019graph} was shown to be stable to permutations of vertex indices, and to small perturbations of edge weights.  \cite{gao:graphScat2018} established the viability of graph scattering coefficients as universal graph features for data analysis tasks (e.g., in social networks and biochemistry data). The wavelets used in \cite{gao:graphScat2018} are similar to those used in \cite{gama:diffScatGraphs2018}, but are constructed from an asymmetric lazy random walk matrix (the wavelets in \cite{gama:diffScatGraphs2018} are constructed from a symmetric matrix). The constructions of \cite{gama:diffScatGraphs2018} and \cite{gao:graphScat2018} were unified in \cite{perlmutter:graphscattering}, which introduced a  large family of graph wavelets, including both  those from  \cite{gama:diffScatGraphs2018} and \cite{gao:graphScat2018} as special cases, and showed that the resulting scattering transforms enjoyed many of the same theoretical properties as in \cite{gama:diffScatGraphs2018}.

In this paper we consider the manifold aspect of geometric deep learning. There are two basic tasks in this setting: (1) classification of multiple signals over a single, fixed manifold; and (2) classification of multiple manifolds. Beyond these two tasks, there are additional problems of interest such as manifold alignment, partial manifold reconstruction, and generative models. Fundamentally for all of these tasks, both in the approach described here and in other papers, one needs to process signals over a manifold. Indeed, even in manifold classification tasks and related problems such as manifold alignment,  one often begins with a set of universal features that can be defined on any manifold, and which are processed in such a way that allows for comparison of two or more manifolds. In order to carry out these tasks, a representation of manifold supported signals needs to be stable to orientations, noise, and deformations over the manifold geometry. Working towards these goals, we define a scattering transform on compact smooth Riemannian manifolds without boundary, which we call \emph{geometric scattering}. Our construction is based on convolutional filters defined spectrally via the eigendecomposition of the Laplace-Beltrami operator over the manifold, as discussed in Section~\ref{sec: geometric wavelet transforms on manifolds}. We show that these convolutional operators can be used to construct a wavelet frame similar to the diffusion wavelets constructed in \cite{coifman:diffWavelets2006}. Then,  in Section~\ref{sec: the geometric scattering transform - big section}, we construct a cascade of these generalized convolutions and pointwise absolute value operations that is used to map signals on the manifold to scattering coefficients that encode approximate local invariance to isometries, which correspond to translations, rotations, and reflections in Euclidean space. We then show in Section \ref{sec: diffeomorphism stability} that our scattering coefficients are also stable to the action of diffeomorphisms with a  notion of stability analogous to the Lipschitz stability considered in~\cite{mallat:scattering2012} on Euclidean space. Our results provide a path forward for utilizing the scattering mathematical framework to analyze and understand geometric deep learning, while also shedding light on the challenges involved in such generalization to non-Euclidean domains. Indeed, while these results are  analogous to those obtained for the Euclidean scattering transform, we emphasize that the underlying mathematical techniques are derived from spectral geometry, which plays no role in the Euclidean analysis. Numerical results in Section \ref{sec: numerics} show that geometric scattering coefficients achieve competitive results for signal classification on a single manifold, and classification of different manifolds. We demonstrate the geometric scattering method can capture the both local and global features to generate useful latent representations for various downstream tasks. Proofs of technical results are provided in the appendices.

\subsection{Notation}

Let $\Mc$ denote a compact, smooth, connected $d$-dimensional Riemannian manifold without boundary contained in $\R^n$, and let  $\Lb^2 (\Mc)$  denote the set of functions $f : \Mc \rightarrow \R$ that are square integrable with respect to the Riemannian volume $dx.$ Let $r(x,x')$ denote the geodesic distance between two points, and let $\Delta$ denote the Laplace-Beltrami operator on $\Mc$. Let $\Isom(\Mc, \Mc')$ denote the set of all isometries between two manifolds $\Mc$ and $\Mc'$, and set $\Isom (\Mc) = \Isom (\Mc, \Mc)$ to be the isometry group of $\Mc$. Likewise, we set $\Diff (\Mc) = \Diff (\Mc, \Mc)$ to be the diffeomorphism group on $\Mc$. For $\zeta \in \Diff (\Mc)$, we let $\| \zeta \|_{\infty} \coloneqq \sup_{x \in \Mc} r(x, \zeta (x))$ denote its maximum displacement.

\section{Geometric wavelet transforms on manifolds}
\label{sec: geometric wavelet transforms on manifolds}

The Euclidean  
scattering transform is constructed using wavelet and low-pass filters defined on $\R^d$. In Section \ref{sec: convolution on manifolds}, we extend the notion of convolution against a filter (wavelet, low-pass, or otherwise), to manifolds using notions from spectral geometry. Many of the notions described in this section are geometric analogues of similar constructions used in graph signal processing \citep{shuman:graphSigProc2013}. Section \ref{sec: frames over manifolds} utilizes these constructions to define Littlewood-Paley frames for $\Lb^2 (\Mc)$, and Section \ref{sec: geometric wavelets} describes a specific class of Littlewood-Paley frames which we call {\it geometric wavelets}. 

\subsection{Convolution on manifolds}
\label{sec: convolution on manifolds}

On $\R^d,$ the convolution of a signal $f\in\Lb^2(\R^d)$ with a filter $h\in\Lb^2(\R^d)$ is defined by translating $h$ against $f$; however, translations are not well-defined on generic manifolds. Nevertheless, convolution can also be characterized using the Fourier convolution theorem, i.e.,  $\widehat{f \ast h}(\omega) = \hf (\omega) \hh (\omega)$.  Fourier analysis   can be defined on $\Mc$ using the spectral decomposition of $-\Delta$. Since $\Mc$ is compact and connected, $-\Delta$ has countably many eigenvalues which we enumerate  as $0 = \lambda_0 < \lambda_1 \leq \lambda_2$ (repeating those with multiplicity greater than one), and there exists a sequence of eigenfunctions $\varphi_0,\varphi_1,\varphi_2,\ldots$ such that $\{ \varphi_k \}_{k \geq 0}$ is an orthonormal basis for $\Lb^2 (\Mc)$ and $-\Delta\varphi_k=\lambda_k\varphi_k.$  While one can take each $\varphi_k$ to be real valued, we do not assume this choice of eigenbasis. One can show that $\varphi_0$ is constant, which implies, by orthogonality, that  $\varphi_k$ has mean zero for $k\geq 1.$ We  consider the eigenfunctions $\{ \varphi_k \}_{k \geq 0}$ as the Fourier modes of the manifold $\Mc$, and define the Fourier series $\hf \in \ellb^2$ of $f \in \Lb^2 (\Mc)$ as
\begin{equation*}
    \hf (k) \coloneqq \langle f, \varphi_k \rangle = \int_{\Mc} f(y) \overline{\varphi_k (y)} \, dy \, .
\end{equation*}

Since $\varphi_0,\varphi_1,\varphi_2,\ldots$ form an orthonormal basis, we have \begin{equation} \label{eqn: fourier inversion}
    f(x) =  \sum_{k \geq 0} \langle f, \varphi_k \rangle \varphi_k(x)= \sum_{k \geq 0} \hf (k) \varphi_k(x) = \sum_{k \geq 0} \left( \int_{\Mc} f(y) \overline{\varphi_k (y)} \, dy \right)  \varphi_k(x) \, .
\end{equation}
For $f, h \in \Lb^2 (\Mc),$ we define the convolution $\ast$ over $\Mc$ between $f$ and $h$ as
\begin{equation} \label{eqn: convolution on M}
    f \ast h (x) \coloneqq \sum_{k \geq 0} \hf (k) \hh (k) \varphi_k (x) \, . 
\end{equation}
We let $T_h : \Lb^2 (\Mc) \rightarrow \Lb^2 (\Mc)$ be the corresponding operator $T_hf(x)\coloneqq f\ast h (x)$ and note that we may write
\begin{equation*}
    T_hf(x)=f \ast h (x) = \int_{\Mc} \left( \sum_{k \geq 0} \hh (k) \varphi_k (x) \overline{\varphi_k (y) }\right) f(y) \, dy = \int_{\Mc} K_h(x,y) f(y) \, dy\, ,
\end{equation*}
where
\begin{equation*}
    K_h (x,y) := \sum_{k \geq 0} \hh (k) \varphi_k (x) \overline{\varphi_k (y)} \, .
\end{equation*}

It is well known that convolution on $\R^d$ commutes with translations. This equivariance property is fundamental to Euclidean ConvNets, and has spurred the development of equivariant  neural networks on other spaces, e.g.,  \cite{pmlr-v48-cohenc16, kondor:equivarianceNNGroups2018, thomas:tensorFieldNetworks2018, kondor:clebsch-gordanNets2018, cohen:sphericalCNNs2018, kondor:covariantCompNets2018, NIPS2018_8239}. Since translations are not well-defined on $\Mc,$ we instead seek to construct a family of operators which commute with isometries. Towards this end, we  say a filter $h \in \Lb^2 (\Mc)$ is a \textit{spectral filter} if $\lambda_k = \lambda_{\ell}$ implies $\hh (k) = \hh (\ell).$ In this case, there exists a function $H:[0,\infty) \rightarrow \R$, which we refer to as the spectral function of $h,$ such that 
\begin{equation*}
H(\lambda_k)=\hh(k) \, , \quad\text{for all }k\geq 0 \, .
\end{equation*}
In the proofs of our theorems, it will be convenient to group together eigenfunctions belonging to the same eigenspace. This motivates us to define 
\begin{equation*}
    \Lambda \coloneqq \{ \lambda \in \R : \text{ there exists } \varphi \in \Lb^2 (\Mc) \text{ such that } -\Delta \varphi = \lambda \varphi \}
\end{equation*}
as the set of all eigenvalues of $-\Delta,$ and to let 
\begin{equation}\label{eqn: Klambda definition}
     K^{(\lambda)}(x,y)\coloneqq\sum_{\lambda_k=\lambda}\varphi_k(x)\overline{\varphi_k(y)}
\end{equation}
for each $\lambda\in\Lambda.$ We  note that if $h$ is a spectral filter, then we may write \begin{equation} \label{eqn: splitupK}
    K_h(x,y) =  \sum_{k \geq 0} H (\lambda_k) \varphi_k (x) \overline{\varphi_k (y) } = \sum_{\lambda\in\Lambda} H(\lambda) K^{(\lambda)}(x,y) \, .    
\end{equation}

For a diffeomorphism $\zeta \in \Diff (\Mc)$ we define the operator $V_{\zeta} : \Lb^2 (\Mc) \rightarrow \Lb^2 (\Mc)$ as
\begin{equation*}
    V_{\zeta} f (x) \coloneqq f (\zeta^{-1} (x)) \, .
\end{equation*}
The operator $V_{\zeta}$ deforms the function $f \in \Lb^2 (\Mc)$ according to the diffeomorphism $\zeta$ of the underlying manifold $\Mc$. The following theorem shows that $T_h$ and $V_\zeta$ commute if $\zeta$ is an isometry and $h$ is a spectral filter. We note the assumption that $h$ is a spectral filter is critical and in general $T_h$ does not commute with isometries if $h$ is not a spectral filter. The proof is in Appendix \ref{sec: pf of isometry equivariance}.

\begin{restatable}[]{thm}{isoequi}
\label{thm: isometry equivariance}
For every spectral filter $h \in \Lb^2 (\Mc)$ and for every $f \in \Lb^2 (\Mc)$, 
\begin{equation*}
    T_h V_{\zeta} f = V_{\zeta} T_h f \, , \quad \forall \, \zeta \in \Isom (\Mc) \, .
\end{equation*}
\end{restatable}

\subsection{Littlewood-Paley frames over manifolds}
\label{sec: frames over manifolds}

A family of spectral filters $\{ h_{\gamma} : \gamma \in \Gamma \}$ (with $\Gamma$ countable), is called a \textit{Littlewood-Paley frame} if it satisfies the following condition which implies that the $h_\gamma$ cover the frequencies of $\Mc$ evenly:
\begin{equation} \label{eqn: lp type frame condition}
     \sum_{\gamma\in\Gamma} |\hh_\gamma (k)|^2 =1 \, , \quad \forall \, k\geq 0 \, .
\end{equation}
We define the corresponding frame analysis operator, $\mathcal{H} : \Lb^2 (\Mc) \rightarrow \ellb^2 (\Lb^2 (\Mc))$, by
\begin{equation*}
    \mathcal{H} f \coloneqq \{ f \ast h_{\gamma} : \gamma \in \Gamma \} \, .
\end{equation*}
The following proposition shows that if \eqref{eqn: lp type frame condition} holds, then $\mathcal{H}f$ preserves the energy of $f$.

\begin{restatable}[]{prop}{lpframe}
\label{prop: lp frame} 
If $\{ h_{\gamma} : \gamma \in \Gamma \}$ satisfies \eqref{eqn: lp type frame condition}, then $\mathcal{H}:\Lb^2 (\Mc)\rightarrow\ellb^2 (\Lb^2 (\Mc))$ is an isometry, i.e.,
\begin{equation*}
     \| \mathcal{H} f \|_{2,2}^2 \coloneqq \sum_{\gamma \in \Gamma} \| f \ast h_{\gamma} \|_2^2 = \| f \|_2^2 \, , \quad \forall \, f \in \Lb^2 (\Mc) \, .
\end{equation*}
\end{restatable}

The proof of Proposition \ref{prop: lp frame} is nearly identical to the corresponding result in the Euclidean case. For the sake of completeness, we provide full details in Appendix \ref{sec: pf of prop: lp frame}. Since the operator $\mathcal{H}$ is linear, Proposition \ref{prop: lp frame} also shows the operator $\mathcal{H}$ is non-expansive, i.e., $\| \mathcal{H} f_1 - \mathcal{H}  f_2\|_{2,2} \leq \| f_1-f_2 \|_2$. This property is directly related to the $\Lb^2$ stability of a ConvNet of the form $\sigma_m (\mathcal{H}_m (\sigma _{m-1} ( \mathcal{H}_{m-1} \cdots \sigma_1 ( \mathcal{H}_1 f )))$, where the $\sigma_{\ell}$ are nonlinear functions. Indeed, if all the frame analysis operators $\mathcal{H}_{\ell}$ and all the nonlinear operators $\sigma_{\ell}$ are  non-expansive, then the entire network is non-expansive as well. 

\subsection{Geometric wavelet transforms on manifolds}
\label{sec: geometric wavelets}

The geometric wavelet transform is a special type of Littlewood-Paley frame analysis operator in which the filters group the frequencies of $\Mc$ into dyadic packets. A spectral filter $\phi \in \Lb^2 (\Mc)$ is said to be a \textit{low-pass filter} if $\hphi (0) = 1$ and $|\hphi (k)|$  is non-increasing with respect to $k$. Typically, $|\hphi (k)|$ decays rapidly as $k$ grows large. Thus, a low-pass filtering,  $T_{\phi} f \coloneqq f \ast \phi$, retains the low frequencies of $f$ while suppressing the high frequencies. A \textit{wavelet} $\psi \in \Lb^2 (\Mc)$ is a spectral filter such that $\hpsi (0) = 0$. Unlike low-pass filters, wavelets have no frequency response at $k = 0,$ but are generally well localized in the frequency domain away from $k=0.$ 

We shall define a family of low-pass and a wavelet filters, using the difference between low-pass filters at consecutive dyadic scales, in a manner which mimics standard wavelet constructions \cite[see, e.g.,][]{meyer:waveletsOperators1993}. Let $G : [0, \infty) \rightarrow \R$ be a non-negative, non-increasing function with $G(0) = 1$. Define a low-pass spectral filter $\phi$ and its dilation at scale $2^j$ for $j \in \Z$ by:
\begin{equation*}
    \hphi (k) \coloneqq G (\lambda_k) \quad \text{and} \quad \hphi_j (k) := G(2^j \lambda_k) \, .
\end{equation*}
Given the dilated low-pass filters, $\{\hphi_j\}_{j \in \mathbb{Z}},$  we defined our wavelet filters by
\begin{equation}\label{eqn: telescope}
    \hpsi_j (k) \coloneqq \Big[ |\hphi_{j-1} (k)|^2 - |\hphi_j (k)|^2 \Big]^{1/2} \, .
\end{equation}

For $J\in\mathbb{Z},$ we let $A_Jf := f \ast \phi_J$ and $\Psi_j f := f \ast \psi_j.$ We then define the \textit{geometric wavelet transform} as 
\begin{equation*}
    \Wc_J f \coloneqq \{ A_J f \, , \, \Psi_j f : j \leq J \} = \{ f \ast \phi_J \, , \, f \ast \psi_j : j\leq J \} \, .
\end{equation*}
The geometric wavelet transform extracts the low frequency, slow transitions of $f$ over $\Mc$ through $A_J f$, and groups the high frequency, sharp transitions of $f$ over $\Mc$ into different dyadic frequency bands via the collection $\{ \Psi_j f : j \leq J \}$. The following proposition can be proved by observing that $\{ \phi_J \, , \, \psi_j : j \leq J \}$ forms a Littlewood-Paley frame and applying Proposition \ref{prop: lp frame}. The proof is nearly identical to the corresponding result in the Euclidean case; however, we provide full details in Appendix \ref{sec: proof pf of prop: wavelet isometry} in order to help keep this paper self-contained.

\begin{restatable}[]{prop}{waveisom}
\label{prop: wavelet isometry}
For any $J \in \Z$,  $\Wc_J:\Lb^2 (\Mc)\rightarrow\ellb^2 (\Lb^2 (\Mc))$ is 
an isometry, i.e.,
\begin{equation*}
    \| \Wc_J f \|_{2,2} = \| f \|_2 \, , \quad \forall \, f \in \Lb^2 (\Mc) \, .
\end{equation*}
\end{restatable}

\begin{figure}
    \centering
    \includegraphics[width=0.90\textwidth]{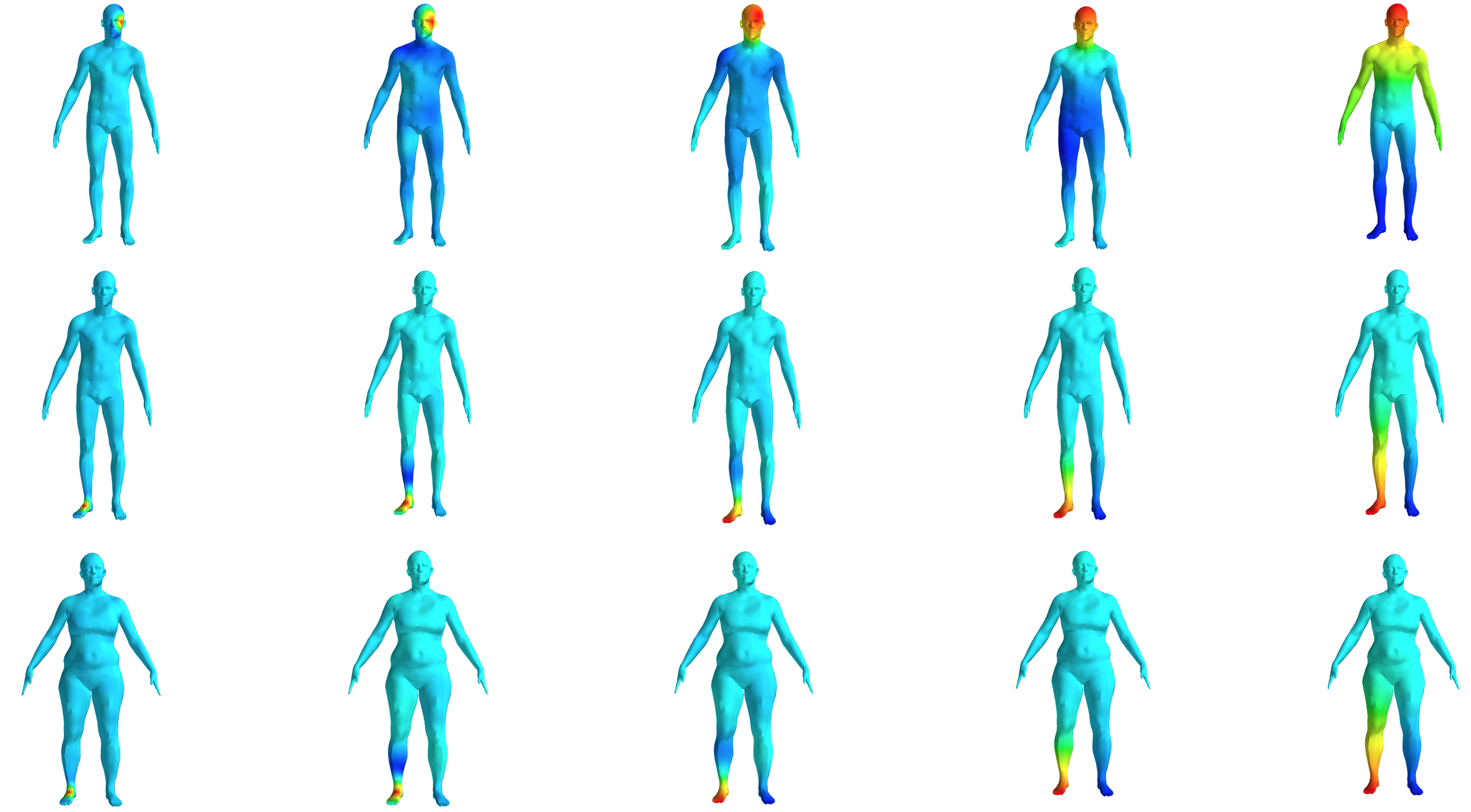}
    \caption{Geometric wavelets on the FAUST mesh with $G(\lambda) = e^{-\lambda}$. From left to right $j = -1, -3, -5, -7, -9$. Positive values are colored red, while negative values are dark blue.}
    \label{fig:geometric wavelets faust}
\end{figure}

An important example  is  $G (\lambda) = e^{-\lambda}$. In this case the low-pass kernel $K_{\phi_J}$ is the heat kernel on $\Mc$ at diffusion time $t = 2^J$, and the wavelet operators $\Psi_j$ are similar to the diffusion wavelets introduced in \cite{coifman:diffWavelets2006}. We also note that wavelet constructions similar to ours were used in \cite{hammond:graphWavelets2011} and \cite{dong2017sparse}. Figure \ref{fig:geometric wavelets faust} depicts these wavelets over manifolds from the FAUST data set \citep{Bogo:CVPR:2014}. Unlike many wavelets commonly used in computer vision, our wavelets are not directional. Indeed, on a generic manifold the concept of directional wavelets is not well-defined since the isometry group cannot be decomposed into translations, rotations, and reflections. Instead, our wavelets have a donut-like shape which is somewhat similar to the wavelet obtained by applying the Laplacian operator on $\R^d$ to a $d$-dimensional Gaussian.

\section{The geometric wavelet scattering transform}
\label{sec: the geometric scattering transform - big section}

The geometric wavelet scattering transform is a type of geometric ConvNet, 
constructed in a manner analogous to the Euclidean scattering transform \citep{mallat:scattering2012} as an alternating cascade of geometric wavelet transforms (defined in Section \ref{sec: geometric wavelets})
 and nonlinearities. As we shall show in Sections \ref{sec: isometric invariance}, \ref{sec: invariance between different manifolds}, and \ref{sec: diffeomorphism stability}, this transformation enjoys several desirable properties for processing data consisting of signals defined on a fixed manifold $\Mc$, 
in addition to tasks in which each data point is a different manifold and one is required to compare and classify manifolds. Tasks of the latter form are approachable due to the use of geometric wavelets that are derived from a universal frequency function $G: [0,\infty) \rightarrow \R$ that is defined independent of $\Mc$. Motivation for these invariance and stability properties is given in Section \ref{sec: role of invariance and stability}, and the geometric wavelet scattering transform is defined in Section \ref{sec: geometeric wavelet scattering operators}. We note that much of our analysis remains valid when our wavelets are replaced with a general Littlewood-Paley frame. However, we will focus on the wavelet case for the ease of exposition and to to emphasize the connections between the manifold scattering transform and its Euclidean analogue.

\subsection{The role of invariance and stability}
\label{sec: role of invariance and stability}

Invariance and stability play a fundamental role in many machine learning tasks, particularly in computer vision. For classification and regression, one often wants to consider two signals $f_1, f_2 \in \Lb^2 (\Mc)$, or two manifolds $\Mc$ and $\Mc'$, to be equivalent if they differ by the action of a global isometry. Similarly, it is desirable that the action of small diffeomorphisms on $f \in \Lb^2 (\Mc)$, or on the underlying manifold $\Mc$, should not have a large impact on the representation of the input signal.

Thus, we seek to construct a family of representations, $(\Theta_t)_{t \in (0, \infty)}$, which are invariant to isometric transformations up to the scale $t$. In the case of analyzing multiple signals on a fixed manifold, such a representation should satisfy a condition of the form: 
\begin{equation} \label{eqn: generic isometry invariance}
    \| \Theta_t (f) - \Theta_t (V_{\zeta} f) \|_{2,2} \leq \alpha (\zeta) \beta (t) \| f \|_2 \, , \quad \forall \, f \in \Lb^2 (\Mc) \, , \, \zeta \in \Isom (\Mc) \, ,
\end{equation}
where $\alpha (\zeta)$ measures the size of the isometry with $\alpha (\id) = 0$, and $\beta(t)$ decreases to zero as the scale $t$ grows to infinity. For diffeomorphisms, invariance is too strong of a property since we are often interested in non-isometric differences between signals on a fixed manifold, or geometric differences between multiple manifolds, not just topological differences, e.g., we often wish to classify a doughnut differently than a coffee mug, even though they are both topologically a $2$-torus. Instead, we want a family of representations that is stable to diffeomorphism actions, but not invariant. Combining this requirement with the isometry invariance condition \eqref{eqn: generic isometry invariance} leads us to seek, for the case of a fixed manifold $\Mc$, a condition of the form:
\begin{equation} \label{eqn: ideal bound}
    \| \Theta_t(f) - \Theta_t (V_{\zeta} f) \|_{2,2} \leq [\alpha (\zeta) \beta (t) + A(\zeta)] \| f \|_2 \, , \, \forall \, t \in (0, \infty) \, , \, f \in \Lb^2 (\Mc) \, , \, \zeta \in \Diff (\Mc) \, ,
\end{equation}
where $A(\zeta)$ measures how much $\zeta$ differs from being an isometry, with $A(\zeta) = 0$ if $\zeta \in \Isom (\Mc)$ and $A(\zeta) > 0$ if $\zeta \notin \Isom (\Mc)$. We also develop analogous conditions for isometries in the case of multiple manifolds in Section \ref{sec: invariance between different manifolds}.

At the same time, the representations $(\Theta_t)_{t \in (0, \infty)}$ should not be trivial. Different classes or types of signals are often distinguished by their high-frequency content, i.e., $\hat{f}(k)$ for large $k.$ The same may also be true of two manifolds $\Mc$ and $\Mc'$, in that differences between them are only readily apparent when comparing the high frequency eigenfunctions of their respective Laplace-Beltrami operators. Our problem is thus to find a family of representations  for data defined on a manifold that is stable to diffeomorphisms, allows one to control the scale of isometric invariance, and discriminates between different types of signals, in both high and low frequencies.  The wavelet scattering transform of \cite{mallat:scattering2012} achieves goals analogous to the ones presented here, but for Euclidean supported signals. We seek to construct a geometric version of the scattering transform, using filters corresponding to the spectral geometry of $\Mc,$ and to show it has similar properties.

\subsection{Defining the geometric wavelet scattering transform}
\label{sec: geometeric wavelet scattering operators}

The \textit{geometric wavelet scattering transform} is a nonlinear operator $S_J : \Lb^2 (\Mc) \rightarrow \ellb^2 (\Lb^2 (\Mc))$ constructed through an alternating cascade of geometric wavelet transforms $\Wc_J$ and nonlinearities. Its construction is motivated by the desire to obtain localized isometry invariance and stability to diffeomorphisms, as formulated in Section \ref{sec: role of invariance and stability}.

A simple way to obtain a locally isometry invariant representation of a signal is to 
apply the low-pass averaging operator $A_J .$ If $G(\lambda)\leq e^{-\lambda},$ then one can use  Theorem \ref{thm: isometry equivariance} 
to show that\begin{equation} \label{eqn: low pass iso invariance}
    \| A_J f - A_J V_{\zeta} f \|_2 %= \| A_J f - V_{\zeta} A_J f \|_2 
    \leq C (\Mc) 2^{-dJ} \| \zeta \|_{\infty} \| f \|_2 \, , \quad \forall \, f \in \Lb^2 (\Mc), \, \forall \zeta \in \Isom(\Mc) \, .
\end{equation}
In other words, the $\Lb^2$ difference between $f \ast \phi_J$ and $V_{\zeta} f \ast \phi_J$ for a unit energy signal $f$ (i.e., $\| f \|_2 = 1$), is no more than the size of the isometry $\| \zeta \|_{\infty}$ depressed by a factor of $2^{dJ}$, up to some universal constant that depends only on $\Mc$. Thus, the parameter $J$ controls the degree of invariance. 

However, by definition $A_J f = f \ast \phi_J = \sum_{k \geq 0} \hf (k) \hphi_J (k) \varphi_k$, and so if $|\hphi_J (k)| \leq e^{-2^J \lambda_k}$, we see the high-frequency content of $f$ is lost in the representation $A_J f$. The high frequencies of $f$ are recovered with the wavelet coefficients $\{ \Psi_j f = f \ast \psi_j : j \leq J \}$, which are guaranteed to capture the remaining frequency content of $f$. However, the wavelet coefficients $\Psi_j f$ are not isometry invariant and thus do not satisfy any bound analogous to \eqref{eqn: low pass iso invariance}. If we apply the averaging operator in addition to the wavelet coefficient operator, we obtain:
\begin{equation*}
    A_J \Psi_j f = f \ast \psi_j \ast \phi_J = \sum_{k \geq 0} \hf (k) \hpsi_j (k) \hphi_J (k) \varphi_k \, ,
\end{equation*}
but by design the sequences $\hphi_J$ and $\hpsi_j$ have small overlapping support in order to satisfy the Littlewood-Paley condition \eqref{eqn: lp type frame condition}, particularly in  their largest responses, and thus $f \ast \psi_j \ast \phi_J \approx 0$. In order to obtain a non-trivial invariant that also retains some of the high-frequency information in the signal $f$, we need to apply a nonlinear operator. Because it is non-expansive and commutes with isometries, we choose the absolute value (complex modulus) function as our non-linearity, and let 
\begin{equation*}
    U_J[j]f\coloneqq | f \ast \psi_j | \, , \quad j \leq J \, .
\end{equation*}
We  then  convolve the $U_J[j]f$ with the low-pass $\phi_j$ to obtain locally invariant descriptions of $f,$  which we refer to as the first-order scattering coefficients:
\begin{equation} \label{eqn: first order scattering}
    S_J [j] f \coloneqq | f \ast \psi_j | \ast \phi_J \, , \quad j \leq J \, .
\end{equation}
The collection of all such coefficients can be  written as 
\begin{equation*}
    A_J U_J^1 f \coloneqq \{ A_JU_J[j]f : j \leq J \}= \{ | f \ast \psi_j | \ast \phi_J : j \leq J \} \, ,
\end{equation*}
where 
\begin{equation*}
    U_J^1 f \coloneqq \{ U_J[j]f : j \leq J \} = \{ |f \ast \psi_j| : j \leq J \} \, .
\end{equation*}
These coefficients satisfy a local invariance bound similar to \eqref{eqn: low pass iso invariance}, but encode multiscale characteristics of $f$ over the manifold geometry, which are not contained in $A_J f$. Nevertheless, the geometric scattering representation $S_J^1 f \coloneqq \{ A_J f \, , \, A_J U_J^1 f \}$ still loses information contained in the signal $f$. Indeed, even with the absolute value, the functions $| f \ast \psi_j|$ have frequency information not captured by the low-pass $\phi_J$. Iterating the geometric wavelet transform $W_J$ recovers this information by computing $W_J U^1_J f = \{ |f \ast \psi_{j_1}| \ast \phi_J \, , \, | f \ast \psi_{j_1} | \ast \psi_{j_2} : j_1, j_2 \leq J \}$, which contains the first order invariants \eqref{eqn: first order scattering} but also retains the high frequencies of $U_J^1 f$. We then obtain second-order geometric wavelet scattering coefficients given by
\begin{equation*}
    S_J [j_1,j_2] f \coloneqq A_JU_J[j_1]U_J[j_2]f=|| f \ast \psi_{j_1}| \ast \psi_{j_2}| \ast \phi_J \, ,
\end{equation*} 
the collection of which can be written as $A_J U_J^1 U_J^1 f$. The corresponding geometric scattering transform up to order $m = 2$ computes $S_J^2 f \coloneqq \{ A_J f \, , \, A_J U_J^1 f \, , \, A_J U_J^1 U_J^1 f \}$, which can be thought of as a three-layer geometric ConvNet that extracts invariant representations of the input signal at each layer. Second order coefficients, in particular, decompose the interference patterns in $| f \ast \psi_{j_1}|$ into dyadic frequency bands via a second wavelet transform. This second order transform has the effect of coupling two scales $2^{j_1}$ and $2^{j_2}$ over the geometry of the manifold $\Mc$. 

\begin{figure}
    \centering
    \adjustbox{width=\textwidth}{% \documentclass[tikz]{standalone}
% \usetikzlibrary{positioning}
% \begin{document}
\begin{tikzpicture}

\draw[dashed,blue,fill=blue!10] (1.85,-0.65) rectangle +(14.2,-1.05);

\node (f) {$f$}; 
\node[draw=black,right=0.5cm of f] (w1) {$W_J$};
\node[draw=black,right=0.5cm of w1] (psi1) {$f\ast\psi_{j_1}$};
\node[double,draw=blue,below=0.5cm of psi1] (phi1) {$f\ast\phi_{J}$};
\node[draw=black,right=0.5cm of psi1] (mod1) {$|\cdot|$};
\node[draw=black,right=0.5cm of mod1] (w2) {$W_J$};
\node[draw=black,right=0.5cm of w2] (psi2) {$|f\ast\psi_{j_1}|\ast\psi_{j_2}$};
\node[double,draw=blue,below=0.5cm of psi2] (phi2) {$|f\ast\psi_{j_1}|\ast\phi_{J}$};
\node[draw=black,right=0.5cm of psi2] (mod2) {$|\cdot|$};
\node[draw=black,right=0.5cm of mod2] (w3) {$W_J$};
\node[draw=black,right=0.5cm of w3] (psi3) {$||f\ast\psi_{j_1}|\ast\psi_{j_2}|\ast\psi_{j_3}$};
\node[double,draw=blue,below=0.5cm of psi3] (phi3) {$||f\ast\psi_{j_1}|\ast\psi_{j_2}|\ast\phi_{J}$};

\node[blue,below=0.5cm of w1] (Sf) {$S_J^2 f = $};
% \draw[->,dashed] (f) -- (Sf);

\draw [->,thick] (f.east) -- (w1.west);
\draw [thick] (w1.east) -- (psi1.west);
\draw [->,blue,thick] (w1.east) to[out=0,in=180] (phi1.west);
\draw [->,thick] (psi1.east) -- (mod1.west);
\draw [->,thick] (mod1.east) -- (w2.west);
\draw [thick] (w2.east) -- (psi2.west);
\draw [->,blue,thick] (w2.east)  to[out=0,in=180] (phi2.west);
\draw [->,thick] (psi2.east) -- (mod2.west);
\draw [->,thick] (mod2.east) -- (w3.west);
\draw [thick] (w3.east) -- (psi3.west);
\draw [->,blue,thick] (w3.east) to[out=0,in=180] (phi3.west);
\draw [dotted,thick] (psi3.east) -- +(0.4,0);

\end{tikzpicture}
% \end{document}
}
    \caption{The geometric wavelet scattering transform $S_J^L$, illustrated for $L=2$.}
    \label{fig:geometric scattering}
\end{figure}

The general geometric scattering transform iterates the wavelet transform and absolute value (complex modulus) operators up to an arbitrary depth. Formally, for $J \in \Z$ and $j_1,\ldots,j_m\leq J$ we let 
\begin{equation*}
    U_J[j_1,\ldots,j_m]\coloneqq U_J[j_m]\ldots U_J[j_1]f= |||f \ast \psi_{j_1}| \ast \psi_{j_2}| \ast \cdots \ast \psi_{j_m}|
\end{equation*}
when $m\geq 1,$
and  we let $U_J[\emptyset]f=f$ when $m=0.$ Likewise, we define 
\begin{equation*}
    S_J[j_1,\ldots,j_m]\coloneqq A_JU_J[j_m]\ldots U_J[j_1]f= |||f \ast \psi_{j_1}| \ast \psi_{j_2}| \ast \cdots \ast \psi_{j_m}|\ast \phi_J \, ,
\end{equation*}
and let $S_J[\emptyset]f = f \ast \phi_J.$ We then consider the maps $U_J :\mathbf{L}^2(\Mc) \rightarrow \ell^2(\mathbf{L}^2(\Mc))$ and $S_J: \mathbf{L}^2(\Mc) \rightarrow \ell^2(\mathbf{L}^2(\Mc))$ given by
\begin{equation*}
    U_Jf = \{ U_J [j_1, \ldots, j_m] f : m\geq 0 \, , \, j_i \leq J \enspace \forall \, 1\leq i \leq m \} \, ,
\end{equation*}
and
\begin{equation*}
    S_Jf = \{ S_J [j_1, \ldots, j_m] f : m\geq 0 \, , \, j_i \leq J \enspace \forall \, 1\leq i \leq m \} \, .
\end{equation*}
An illustration of the map $S_J,$ which we refer to as the geometric scattering transform at scale $2^J,$ is given by Figure \ref{fig:geometric scattering}. In practice, one only uses finitely many layers, which motives us to also consider the $L$-layer versions of $U_J$ and $S_J$ defined for $L\geq 0$ by 
\begin{align*}
    U_J^L f  :=& \{ U_J [j_1, \ldots, j_m] f : 0 \leq m \leq L \, , \, j_i \leq J \enspace \forall \, 1\leq i \leq m \}
\end{align*}
and 
\begin{equation*} \label{eqn: definition of finite layer scattering}
    S_J^L f  := \{ S_J [j_1, \ldots, j_m] f : 0 \leq m \leq L \, , \, j_i \leq J \enspace \forall \, 1\leq i \leq m \} \, .
\end{equation*}
The invariance properties of $S_J$ and $S_J^L$ are described in Sections \ref{sec: isometric invariance} and \ref{sec: invariance between different manifolds}, whereas their diffeomorphism stability properties are described in Section \ref{sec: diffeomorphism stability}. The following proposition shows that both $S_J$ and $S_J^L$ are non-expansive. 

\begin{prop}\label{prop: nonexpansive}
Both the finite-layer and infinite-layer geometric wavelet scattering transforms are nonexpansive. Specifically, 
\begin{equation*}
\| S^L_J f_1 - S^L_J f_2 \|_{2,2}\leq \| S_J f_1 - S_J f_2 \|_{2,2} \leq \| f_1 - f_2 \| \, , \quad \forall \, f_1, f_2 \in \Lb^2 (\Mc) \, .
\end{equation*}
\end{prop}

The first inequality is trivial. The proof of the second inequality is nearly identical to \citet[Proposition 2.5]{mallat:scattering2012}, and is thus omitted.

\subsection{Isometric invariance}
\label{sec: isometric invariance}

The geometric wavelet scattering transform is invariant to the action of the isometry group on the input signal $f$ up to a factor that depends upon the frequency decay of the low-pass spectral filter $\phi_J$. In particular, the following theorem establishes isometric invariance up to the scale $2^{J}$ under the assumption that $\hphi (k) = G(\lambda_k) \leq e^{-\lambda_k}.$ The proof of Theorem \ref{thm: isometric invariance} is given in Appendix \ref{sec: proof of isometry invariance}. 

\begin{restatable}[]{thm}{isoinv}
\label{thm: isometric invariance}
Let $\zeta \in \Isom (\Mc)$ and suppose $G(\lambda)\leq e^{-\lambda}$. Then there is a constant $C (\Mc) < \infty$ such that for all $f \in \Lb^2 (\Mc),$
\begin{equation} \label{eqn: isometric invariance for infinite depth network}
    \| S_J f - S_J V_{\zeta} f \|_{2,2} \leq C ( \Mc )  2^{-dJ} \| \zeta \|_{\infty} \| U_J f \|_{2,2} \, ,
\end{equation}
and 
\begin{equation}\label{eqn: isometric invariance for finite depth network}
    \| S_J^L f - S_J^L V_{\zeta} f \|_{2,2} \leq C ( \Mc ) (L+1)^{1/2} 2^{-dJ} \| \zeta \|_{\infty} \| f \|_2 \, .
\end{equation}
\end{restatable}

The factor $\| U_J f \|_{2,2}$ for an infinite depth network is hard to bound in terms of $\| f \|_2$, which is also true for the Euclidean scattering transform \citep{mallat:scattering2012}. However, for finite depth networks, a simple argument shows that $\| U_J^L f \|_{2,2} \leq (L+1)^{1/2} \| f \|_2$, which yields \eqref{eqn: isometric invariance for finite depth network}.

For manifold classification (or any task requiring rigid invariance), we take $J \rightarrow \infty$. This limit is equivalent to replacing the low-pass operator $A_J$ with an integration over $\Mc$,
\begin{align} 
    \lim_{J \rightarrow \infty} S_J [j_1, \ldots, j_m] f (x) &= \frac{1}{\sqrt{\mathrm{vol}(\Mc)}} \int_{\Mc} |||f \ast \psi_{j_1}| \ast \psi_{j_2}| \ast \cdots \ast \psi_{j_m} (x')| \, dx' \nonumber\\
    &= \mathrm{vol}(\Mc)^{-1/2} \| || f \ast \psi_{j_1} | \ast \psi_{j_2}| \ast \cdots \ast \psi_{j_m} \|_1, \, \label{eqn: J infinity}
\end{align}
where the above limit is in $\mathbf{L}^2(\mathcal{M})$.

Equation \eqref{eqn: J infinity} motivates the definition of a non-windowed geometric scattering transform,
\begin{align*}
    \Sbar f &:= \{ \Sbar f (j_1, \ldots, j_m) : m \geq 0 \, , \, j_i \in \Z \enspace \forall \, 1 \leq i \leq m \} \\
    \Sbar f (j_1, \ldots, j_m) &:= \| || f \ast \psi_{j_1} | \ast \psi_{j_2} | \ast \cdots \ast \psi_{j_m} \|_1 \, .
\end{align*}
We also define $\Sbar^L f$ as the $L$-layer version of $\Sbar f$, analogous to $S_J^L f$ defined in \eqref{eqn: definition of finite layer scattering}. Unlike $Sf \in \ellb^2 (\Lb^2 (\Mc))$, which consists of a countable collection of functions, $\Sbar f$ consists of a countable collections of scalar values. Theorem \ref{thm: isometric invariance} and \eqref{eqn: J infinity} show that these values are invariant to global isometries $\zeta \in \Isom (\Mc)$ acting on $f$. The following proposition shows they form a sequence in $\ellb^2$. We give a proof in Appendix \ref{sec: Proof of Proposition littleelltwo}.
\begin{restatable}[]{thm}{littleelltwo}
\label{prop: littleelltwo}
If $f \in \Lb^2 (\Mc)$, then $\Sbar f \in \ellb^2$ with $\| \Sbar f \|_2 \leq \| f \|_2$.
\end{restatable}

\subsection{Isometric invariance between different manifolds}
\label{sec: invariance between different manifolds}

Let $\Mc$ and $\Mc'$ be isometric manifolds. For shape matching tasks in which $\Mc$ and $\Mc'$ should be identified as the same shape, it is appropriate to let $J \rightarrow \infty,$ and, inspired by \eqref{eqn: J infinity}, use the $\Sbar$ representation to carry out the computation; see Section \ref{sec: faust} for numerical results along these lines. In such tasks, one selects a signal $f$  that it is defined intrinsically in terms of the geometry $\Mc$, i.e., one that is chosen in such a way that given $f$ on $\Mc$,  one can  compute a corresponding signal $f'=V_{\zeta} f$ on $\Mc'$ without explicit knowledge of $\zeta \in \Isom (\Mc, \Mc')$. For example, if $\Mc$ is a two-dimensional surface embedded in $\R^3$, and $\Mc'$ is a three-dimensional rotation of $\Mc$ by $\zeta \in \mathrm{SO}(3)$, then the coordinate function is such a function. Indeed, let $x = (x_1, x_2, x_3) \in \Mc \subset \R^3$ and suppose $f(x) = x_i$ is the coordinate function on $\Mc$ for a fixed coordinate $1 \leq i \leq 3$. Then the coordinate function $f'(x') = x_i'$ on $\Mc'$ is given by $f' = V_{\zeta} f$. More sophisticated examples include the SHOT features of \cite{tombari2010unique, bshot_iros2015}. The following proposition shows that the geometric scattering transform produces a representation  that is invariant to isometries $\zeta \in \Isom (\Mc, \Mc')$. We give a proof in Appendix \ref{Sec: The proof of Prop: nonwindowedinvdiff}.

\begin{restatable}[]{prop}{nonwindowinvariancedifferent} \label{prop: nonwindowedinvdiff}
Let $\zeta \in \Isom (\Mc, \Mc')$, let $f \in \Lb^2 (\Mc),$ and let $f'\coloneqq V_{\zeta} f$ be the corresponding signal defined on $\Mc'.$ Then $\Sbar f = \Sbar' f'$.
\end{restatable}

In other tasks, one may wish to have local isometric invariance between $\Mc$ and $\Mc'$. We thus extend Theorem \ref{thm: isometric invariance} in the following way. If $\zeta_1 \in \Isom (\Mc, \Mc')$, then the operator $V_{\zeta_1}$ maps $\Lb^2(\Mc)$ into $\Lb^2(\Mc').$  We wish to estimate how much $\left(S_J\right)'V_\zeta f$ differs from $S_Jf,$ where $\left(S_J\right)'$ denotes the geometric wavelet scattering transform on $\Mc'.$ However, the difference $S_J f-\left(S_J\right)'V_\zeta f$ is not well-defined since $S_J f$ is a countable collection of functions defined on $\Mc$ and  $\left(S_J\right)'V_\zeta f$ is a collection of functions defined on $\Mc'.$ Therefore, in Theorem~\ref{thm: isoinvariancediff} we let $\zeta_2$ be a second isometry from $\Mc$ to $\Mc'$ and estimate $\|S_Jf-V_{\zeta_2^{-1}}\left(S_J\right)'V_{\zeta_1}f\|_{2,2}$; see Appendix \ref{sec: proof for isometrydiff} for the proof.

\begin{restatable}[]{thm}{isoinvariancediff}
\label{thm: isoinvariancediff}
Let $\zeta_1,\zeta_2 \in \Isom (\Mc, \Mc')$ and assume that $G(\lambda)\leq e^{-\lambda}.$ Then there is a constant $C (\Mc) < \infty$ such that
\begin{equation}\label{eqn: isometric invariance different Infinite}
    \| S_J  f - V_{\zeta_2^{-1}}\left(S_J\right)'  V_{\zeta_1} f \|_{2,2} \leq  C(\Mc)  2^{-dJ} \| \zeta_2^{-1}\circ \zeta_1 \|_{\infty} \| U_J f \|_2 \, , \quad \forall \, f \in \Lb^2(\Mc) \, .
\end{equation}
and
\begin{equation}\label{eqn: isometric invariance different finite}
    \| S^L_J  f - V_{\zeta_2^{-1}}\left(S^L_J\right)'  V_{\zeta_1} f \|_{2,2} \leq  C(\Mc) (L+1)^{1/2} 2^{-dJ} \| \zeta_2^{-1}\circ \zeta_1 \|_{\infty} \| f \|_2 \, , \quad \forall \, f \in \Lb^2(\Mc) \, .
\end{equation}
\end{restatable}
It is worthwhile to contrast Proposition \ref{prop: nonwindowedinvdiff}  and Theorem \ref{thm: isoinvariancediff} with Theorem \ref{thm: isometric invariance} stated in Section \ref{sec: isometric invariance}. As mentioned in the introduction, two of the basic tasks we condsider are the classification of multiple signals over a single, fixed manifold and the classification of multiple manifolds. Since Theorem \ref{thm: isometric invariance} considers an isometry $\zeta:\mathcal{M}\rightarrow\mathcal{M}$, it shows that the manifold scattering transform is well-suited for the former task. Proposition \ref{prop: nonwindowedinvdiff}  and Theorem \ref{thm: isoinvariancediff}, on the other hand, assume that $\zeta$ is an isometry from one manifold $\mathcal{M}$ to another manifold $\mathcal{M}'$, and therefore indicate that the manifold scattering transform is well-suited to the latter task as well.

\section{Stability to Diffeomorphisms}
\label{sec: diffeomorphism stability}

In this section we show the scattering transform is stable to the action of diffeomorphisms on a signal $f \in \Lb^2 (\Mc)$.
% diffeomorphic perturbations. 
In Section \ref{sec: stabilitybl}, we show that when restricted to bandlimited functions, the geometric scattering transform is stable to diffeomorphisms. In Section \ref{sec: single filter stability}, we show that under certain assumptions on $\Mc,$ that spectral filters are stable to diffeomorphisms (even if $f$ is not bandlimited). As a consequence, it follows that finite width (i.e., a finite number of wavelets per layer) scattering networks are stable to diffeomorphisms on these manifolds.

\subsection{Stability for bandlimited functions}
\label{sec: stabilitybl}

Analogously to the Lipschitz diffeomorphism stability in \citet[Section 2.5]{mallat:scattering2012}, we wish to show the geometric scattering coefficients are stable to diffeomorphisms that are close to being an isometry. Similarly to \cite{wiatowski:frameScat2015, czaja:timeFreqScat2017}, we will assume the input signal $f$ is $\lambda$- bandlimited for some $\lambda>0.$ That is, $\hf (k) = \langle f,\varphi_k\rangle=0$ whenever $\lambda_k>\lambda.$ 

\begin{restatable}[]{thm}{smudgestabilityBL}
\label{thm: smudgestablityBL}
Let $\zeta \in \Diff (\Mc),$ and assume $G(\lambda)\leq e^{-\lambda}.$
Then there is a constant $C (\Mc) <\infty$ such that  if $\zeta=\zeta_1\circ\zeta_2$ for some isometry $\zeta_1 \in \Isom (\Mc)$ and diffeomorphism $\zeta_2 \in \Diff (\Mc),$ then 
\begin{equation} \label{eqn: smudgestabilityinfinity}
    \|S_J f - S_J V_\zeta f \|_{2,2} \leq C (\Mc) \Big [  2^{-dJ} \| \zeta_1 \|_{\infty} \|U_J f\|_2 + \lambda^{d} \|\zeta_2\|_\infty \|f\|_2 \Big]  \, ,
\end{equation}
and 
\begin{equation} \label{eqn: smudgestabilityfinite}
    \|S_J^L f - S_J^L V_\zeta f \|_{2,2} \leq C (\Mc) \Big [ (L+1)^{1/2} 2^{-dJ} \| \zeta_1 \|_{\infty} + \lambda^{d} \|\zeta_2\|_\infty \Big] \|f\|_2 \, ,
\end{equation}
for all functions $f \in \Lb^2 (\Mc)$ such that $\hf (k) = \langle f,\varphi_k\rangle=0$ whenever $\lambda_k>\lambda.$
\end{restatable}

Theorem \ref{thm: smudgestablityBL} achieves the goal set forth by \eqref{eqn: ideal bound}, with two exceptions: (i) we restrict to bandlimited functions; and (ii) the infinite depth network has the term $\| U_J f \|_{2,2}$ in the upper bound. We leave the vast majority of the work in resolving these issues to future work, although Section \ref{sec: single filter stability} takes some initial steps in resolving (i). We also leave for future work the case of quantifying $\| \Sbar f - \Sbar'f \|_2$ for two diffeomorphic manifolds $\Mc$ and $\Mc'$. When $\zeta$ is an isometry, it reduces to Theorem \ref{thm: isometric invariance}, since in this case we may choose $\zeta = \zeta_1$, $\zeta_2 = \id$ and note that $\| \id \|_{\infty} = 0$. For a general diffeomorphism $\zeta$, taking the infimum of $\|\zeta_2\|_\infty$ over all factorizations $\zeta = \zeta_1 \circ \zeta_2$ leads to a bound where the first term depends on the scale of the isometric invariance and the second term depends on the distance from $\zeta$ to the isometry group $\Isom (\Mc)$ in the uniform norm. Letting $J\rightarrow \infty,$ we may also prove an analogous theorem for the non-windowed scattering transform.

\begin{restatable}[]{thm}{smudgestabilityBLnowindow}
\label{thm: smudgestablityBLnowindow}
Let $\zeta \in \Diff (\Mc),$ and assume $G(\lambda)\leq e^{-\lambda}.$ Then there is a constant $C (\Mc) <\infty$ such that if $\zeta=\zeta_1\circ\zeta_2$ for some isometry $\zeta_1 \in \Isom (\Mc)$ and diffeomorphism $\zeta_2 \in \Diff (\Mc),$ then 
\begin{equation} 
    \|\overline{S} f - \overline{S} V_\zeta f \|_2 \leq  \lambda^{d} \|\zeta_2\|_\infty \|f\|_2 \, ,
\end{equation}
for all functions $f \in \Lb^2 (\Mc)$ such that $\hf (k) = \langle f,\varphi_k\rangle=0$ whenever $\lambda_k>\lambda.$
\end{restatable}

The proofs of Theorems \ref{thm: smudgestablityBL} and \ref{thm: smudgestablityBLnowindow} are given in Appendix \ref{sec: proof of blstability}.

\subsection{Single-Filter Stability}\label{sec: single filter stability}

Theorems \ref{thm: smudgestablityBL} and \ref{thm: smudgestablityBLnowindow} prove diffeomorphism stability for the geometric wavelet scattering transform, but their proof techniques rely on $f$ being bandlimited. In this section we discuss a possible approach to proving a stability result for all $f \in \Lb^2 (\Mc)$. 

As stated in Theorem \ref{thm: isometry equivariance}, spectral integral operators are equivariant to the action of isometries. This fact is crucial to proving Theorem \ref{thm: isometric invariance} because it allows us to estimate
\begin{equation}\label{left}
    \| S_J f -  V_{\zeta} S_J f \|_{2,2}
\end{equation}
instead of 
\begin{equation}\label{right}
    \| S_J f - S_J V_{\zeta} f \|_{2,2} \, .
\end{equation}
In \cite{mallat:scattering2012}, it is shown that the Euclidean scattering transform $S_{Euc}$ is stable to the action of certain diffeomorphisms which are close to being translations. A key step in the proof is a bound on the commutator norm $\| [S_{Euc}, V_{\zeta}] \|$, which then allows the author to bound a quantity analogous to (\ref{left}) instead of bounding (\ref{right}) directly. 
This motivates us to study the commutator of spectral integral operators with $V_\zeta$ for diffeomorphisms which are close to being isometries. 

For technical reasons, we will assume that $\Mc$ is two-point homogeneous, that is, for any two pairs of points, $(x_1,x_2), ~(y_1,y_2)$ such that $r(x_1,x_2)=r(y_1,y_2),$ there exists an isometry $\zeta \in \Isom (\Mc)$ such that $\zeta(x_1)=y_1$ and $\zeta(x_2)=y_2.$ In order to quantify how far a diffeomorphism $\zeta \in \Diff (\Mc)$ differs from being an isometry we will consider two quantities:
\begin{equation}\label{a1}
    A_1 (\zeta) = \sup_{\substack{x,y \in \Mc \\ x \neq y}} \left|\frac{r\left(\zeta (x), \zeta (y)\right) - r(x,y)}{r(x,y)}\right|,
\end{equation}
and
\begin{equation}\label{a2}
    A_2(\zeta) = \left(\sup_{x \in \Mc} \Big||\det[D\zeta (x)]| - 1\Big|\right)\left(\sup_{x\in\Mc} \Big|\det[D\zeta^{-1}(x)]\Big|\right).
\end{equation}
Intuitively, $A_1$ is a measure of how much $\zeta$ distorts distances, and $A_2$ is a measure of how much $\zeta$ distorts volumes. We let $A(\zeta)=\max\{A_1(\zeta),A_2(\zeta)\}$ and note that if $\zeta$ is an isometry, then $A (\zeta)= 0$. We remark that $A(\zeta)$ defined here differs from the notion of diffeomorphism size used in Theorems \ref{thm: smudgestablityBL} and \ref{thm: smudgestablityBLnowindow}. It is an interesting research direction to understand the differences between these formulations, and to understand more generally which definitions of diffeomorphism size geometric deep networks are stable to. The following theorem, which is proved in Appendix \ref{sec: commutator proof}, bounds the operator norm of $[T_{h}, V_{\zeta}]$ in terms of $A (\zeta)$ and a quantity depending upon $h$.

\begin{restatable}[]{thm}{twopt}
\label{twoptho}
Assume that $\Mc$ is two-point homogeneous, and let $h\in\mathbf{L}^2(\Mc)$ be a spectral filter. Then there exists a constant $C (\Mc) > 0$ such that for any diffeomorphism $\zeta \in \Diff (\Mc)$, 
\begin{align*}
    \| [T_{h}, V_{\zeta}]\| &\leq C (\Mc) A (\zeta) B (h)
\end{align*}
where
\begin{equation*}
    B (h) = \max \left\{ \sum_{k \in \N} \widehat{h}(k) \lambda_k^{(d+1)/4}, \left( \sum_{k \in \N} \widehat{h}(k)^2 \right)^{\frac{1}{2}} \right\}.
\end{equation*}
\end{restatable}

Theorem \ref{twoptho} leads to the following corollary, which we prove in Appendix \ref{sec: The Proof of Corollary  single wavelet stabilty}

\begin{restatable}[]{cor}{singlewavelet}
\label{cor: single wavelet stabilty}
Assume that $\Mc$ is two-point homogeneous and that $G(\lambda)\leq e^{-\lambda},$ then
\begin{align*}
\| [\Psi_j, V_{\zeta}]\| &\leq C (\Mc) A (\zeta) B (\psi_j)
\end{align*}
where,
\begin{equation*}
B (\psi_j) = \max \left\{ 2^{-(d+1/2)(j-1)}, 2^{-dj/4} \right\}.
\end{equation*}
\end{restatable}

In practice, the wavelet transform is implemented using finitely many wavelets. By the triangle inequality, Corollary \ref{cor: single wavelet stabilty} leads to a commutator estimate for the finite wavelet transform. Therefore,  by the arguments used in the proof of Theorem 2.12 in \cite{mallat:scattering2012}, it follows that the geometric scattering transform is stable to diffeomorphisms  on two-point homogeneous manifolds when implemented with finitely many wavelets at each layer. We do note however, that $B(\psi_j)$ increases exponentially as $j$ decreases to $-\infty.$ Therefore, this argument only applies to a finite-wavelet implementation of the geometric scattering transform. \cite{mallat:scattering2012} overcomes this difficulty using an almost orthogonality argument. In the future, one might seek to adapt these techniques to the manifold setting. However, there are numerous technical difficulties which are not present in the Euclidean setting.

\section{Numerical results}
\label{sec: numerics}

In this section, we describe two numerical experiments to illustrate the utility of the geometric wavelet scattering transform. We consider both traditional geometric learning tasks, in which we compare to other geometric deep learning methods, as well as limited training tasks in which the unsupervised nature of the transform is particularly useful. In the former set of tasks, empirical results are not state-of-the-art, but they show that the geometric scattering model is a good mathematical model for geometric deep learning. Specifically, in Section  \ref{sec: mnist} we classify signals, corresponding to digits, on a fixed manifold, the two-dimensional sphere. Then, in Section \ref{sec: faust} we classify different manifolds which correspond to ten different people whose bodies are positioned in ten different ways. The back-end classifier for all experiments is an RBF kernel SVM. 

In order to carry out our numerical experiments, it was necessary to discretize our manifolds and represent them as graphs. We use triangle meshes for all manifolds in this paper, which allows us to approximate the Laplace-Beltrami operator and integration on each manifold via the approach described in \cite{solomon:LBswissKnife2014}. We emaphasize that this approximation of the Laplace-Beletrami operator is not the standard graph Laplacian of the triangular mesh, and thus the discretized geometric scattering transform is not the the same as the versions of the graph scattering transform reported in \cite{zou2019graph, gama:diffScatGraphs2018, gama:stabilityGraphScat2019, gao:graphScat2018}.

\subsection{Spherical MNIST}
\label{sec: mnist}

\begin{figure}
    \subfigure[Non-rotated spherical MNIST classification using $\lim_{J \rightarrow \infty} S_J^L f$ for different network depths $L$. Depth $L=3$ obtains $93$\% classification accuracy.]{
        \makebox[0.435\textwidth]{\includegraphics[width=0.38\linewidth, trim = 0 50 0 0, clip]{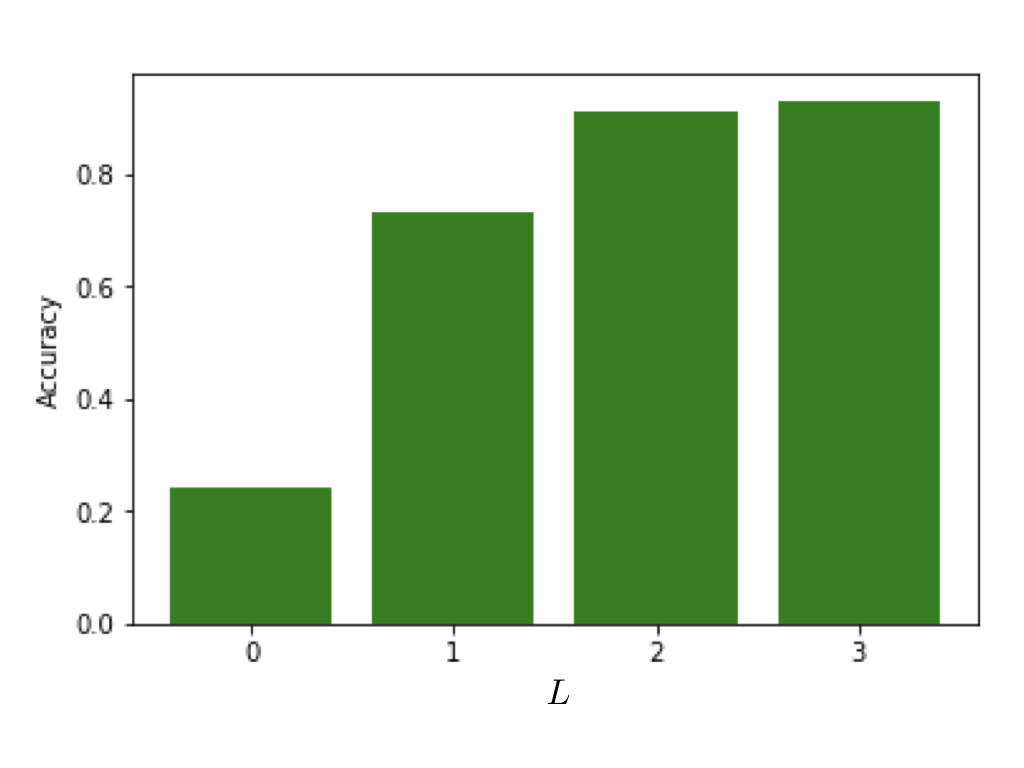}}
        \label{table: scattering order J inf}
    }
    \hfill
    \raisebox{1pt}{
    \subfigure[Spherical MNIST classification with not rotated (NR) and rotated (R) datasets. Note that \cite{cohen:sphericalCNNs2018, kondor:clebsch-gordanNets2018, jiang2018spherical} utilize fully learned filters specifically designed for the sphere.]{
            \makebox[0.515\textwidth]{\adjustbox{width=0.5\linewidth}{
            \begin{tabular}{r|c|c|c|l}
            \hhline{~---~}
            ~ & Model & NR & R \\
            \hhline{~---~}
            & S2CNN \citep{cohen:sphericalCNNs2018} & $0.96$ & $0.95$\\
            \hhline{~---~}
            & FFS2CNN \citep{kondor:clebsch-gordanNets2018} & $0.96$ & $0.97$\\
            \hhline{~---~}
            & Method from \cite{jiang2018spherical} & $0.99$ & N/A \\
            \hhline{~---~}
            & Haar wavelet scattering \citep{chen:scatHaar2014} & $0.90$ & N/A \\
            \hhline{~---~}
            & Geometric scattering, $J = -2$, $L = 2$ &$0.95$ & $0.95$\\
            \hhline{~---~}
            \end{tabular}
            }}
            \label{table: r/r ur/ur mnist}
    }
    }
    \caption{Spherical MNIST classificaion results.}
\end{figure}

In the first experiment, we project the MNIST dataset from Euclidean space onto a two-dimensional sphere using a triangle mesh with 642 vertices. During the projection, we generate two datasets consisting of not rotated (NR) and randomly rotated (R) digits. Using the NR spherical MNIST database, we first investigate in Figure \ref{table: scattering order J inf} the power of the globally invariant wavelet scattering coefficients for different network depths $L$ and with $J \rightarrow \infty$, which is equivalent to using the $\Sbar^L f$ representation defined in Section \ref{sec: isometric invariance}. Here $f$ is the projection of the digit onto the sphere. We observe increasing accuracy but with diminishing returns across the range $0 \leq L \leq 3$. Then on both the NR and R spherical MNIST datasets, we calculate the geometric scattering coefficients $S_J^L f$ for $J = -2$ and $L = 2$. Other values of $J$ are also reported in Appendix \ref{sec: numerical details}, in addition to further details on how the spherical MNIST classification experiments were conducted. From Theorem \ref{thm: smudgestablityBL}, we know the scattering transform is stable to randomly generated rotations and Table \ref{table: r/r ur/ur mnist} shows the scattering coefficients capture enough rotational information to correctly classify the digits.

\subsection{FAUST}
\label{sec: faust}

The FAUST dataset \citep{Bogo:CVPR:2014} contains ten poses from ten people resulting in a total of 100 manifolds represented by triangle meshes. We first consider the problem of classifying poses. This task requires globally invariant features, and thus we compute the globally invariant geometric wavelet scattering transform $\Sbar^L f$ of Section \ref{sec: isometric invariance}. Following the common practice of other geometric deep learning methods \citep[see, e.g.,][]{Litany2017, Lim2018}, we use 352 SHOT features \citep{tombari2010unique, bshot_iros2015} as initial node features $f$. We used 5-fold cross validation for the classification tests with nested cross validation to tune the hyper-parameters of the RBF kernel SVM, as well as the network depth $L$. We remark that tuning the network depth of the geometric scattering transform is relatively simple as compared to fully learned geometric deep networks, since the filters are predefined geometric wavelets. This is particularly important for smaller data sets such as FAUST where there is a limited amount of training data. As indicated in Table 3, we achieve 95\% overall accuracy using the geometric scattering features, compared to 92\% accuracy achieved using only the integrals of SHOT features (i.e., restricting to $L=0$). We note that \cite{DBLP:MasciBBV15} also considered pose classification, but the authors used a different training/test split (50\% for training and 50\% for test in a leave-one-out fashion), so our results are not directly comparable.

As a second task, we attempt to classify the people. This task is even more challenging than classifying the poses since some of the people are very similar to each other. We again performed 5-fold cross-validation, with each fold containing two poses from each person to ensure the folds are evenly distributed. As shown in Table \ref{table: faust classification}, we achieved 76\% accuracy on this task compared to the 61\% accuracy using only integrals of SHOT features. In order to further emphasize the difference between the discretized geometric scattering transform and the graph scattering transform, we also attempted this task using the graph scattering transform derived from the graph Laplacian of the manifold mesh, applied to the SHOT features of each manifold. For this approach, which is representative of the aforementioned graph scattering papers, we obtained 58\% accuracy. This result is similar to the 61\% accuracy obtained by the baseline SHOT feature approach and empirically indicates the importance of encoding geometric information into the scattering transform for manifold-based tasks. More details regarding both tasks are in Appendix \ref{sec: numerical details}. 

\begin{table*}[!htb]
\caption{Manifold classification on FAUST dataset with two tasks.}
\vspace{5pt}
\centering
\adjustbox{width=0.65\linewidth}{\begin{tabular}{r|c|c|c|l}
\hhline{~---~}
~ & Task/Model &SHOT only & Geometric scattering \\
\hhline{~---~}
&Pose classification &$0.92$ &$0.95$\\
\hhline{~---~}
& Person classification&$0.61$ & $0.76$\\
\hhline{~---~}
\end{tabular}}
\label{table: faust classification}
\end{table*}

\section{Conclusion}
We have constructed a geometric version of the scattering transform on a large class of Riemannian manifolds and shown this transform is non-expansive, invariant to isometries, and stable to diffeomorphisms. Our construction uses the spectral decomposition of the Laplace-Beltrami operator to construct a class of spectral filtering operators that generalize convolution on Euclidean space. While our numerical examples demonstrate geometric scattering on two-%(or three-)
dimensional manifolds, our theory remains valid for manifolds of any dimension $d,$ and therefore can be naturally extended and applied to higher-dimensional manifolds in future work. Finally, our construction provides a mathematical framework that enables future analysis and understanding of geometric deep learning.

% Acknowledgments---Will not appear in anonymized version
\acks{This research was partially funded by: grant P42 ES004911, through the National Institute of Environmental Health Sciences of the NIH, supporting \emph{F.G.}; IVADO (l'institut de valorisation des donn\'{e}es) [\emph{G.W.}]; the Alfred P. Sloan Fellowship (grant FG-2016-6607), the DARPA Young Faculty Award (grant D16AP00117), the NSF CAREER award (grant 1845856), and NSF grant 1620216 [\emph{M.H.}]; NIH grant R01GM135929 [\textit{M.H. \& G.W}]. The content provided here is solely the responsibility of the authors and does not necessarily represent the official views of the funding agencies.}

\bibliography{Update}
% \bibliographystyle{unsrt}

%%% NOTICE: when uncommenting the following line for one-file mode, the supplement external dependency line in the preamble has to be commented out !!!
\appendix

\section{Proof of Theorem \ref{thm: isometry equivariance}}
\label{sec: pf of isometry equivariance}

\isoequi*

We will prove a result that generalizes  Theorem \ref{thm: isometry equivariance} to isometries between different manifolds. This more general result will be needed to prove Theorem \ref{thm: isoinvariancediff}.
 
Before stating our more general result, we introduce some notation. Let $\Mc$ and $\Mc'$ be smooth compact connected Riemannian manifolds without boundary, and let $\zeta:\Mc\rightarrow\Mc'$ be an isometry. Since $\Mc$ and $\Mc'$ are and  isometric, their Laplace Beltrami operators  $\Delta$ and $\Delta'$ have the same eigenvalues, and we enumerate the eigenvalues of $-\Delta$ (and also of $-\Delta'$) in increasing order (repeating those with multiplicity greater than one) as $0=\lambda_0<\lambda_1\leq\lambda_2\leq\ldots.$ Recall that if $h\in \Lb^2(\Mc)$ is a spectral filter, then by definition, there exists a function $H:[0,\infty) \rightarrow \R$ such that 
\begin{equation*}
     H(\lambda_k)\coloneqq\widehat{h}(k),\:\text{for all } k\geq 0 \, ,
\end{equation*} 
and that 
\begin{equation*}
    K_h (x,y) = \sum_{k \geq 0} \hh (k) \varphi_k (x) \overline{\varphi_k (y)} = \sum_{k \geq 0} H (\lambda_k) \varphi_k (x) \overline{\varphi_k (y)} \, .
\end{equation*}
Therefore, we may define an operator $T_h',$ on $\Lb^2(\Mc'),$ which we consider the analogue of $T_h,$ as integration against the kernel
\begin{equation*}
    K'_h (x,y) \coloneqq \sum_{k \geq 0} H (\lambda_k) \varphi_k' (x) \overline{\varphi_k' (y)} \, ,
\end{equation*}
where $\varphi'_0,\varphi'_1,\ldots,$ is an orthonormal basis of eigenfunction on $\Lb^2(\Mc')$ with $-\Delta'\varphi_k'=\lambda_k\varphi_k'.$ With this notation, we may now state a generalized version of Theorem \ref{thm: isometry equivariance} (to recover Theorem \ref{thm: isometry equivariance}, we set $\Mc'=\Mc$).
 
\begin{thm}\label{thm: equivariance, different mfolds} 
Let $\zeta:\Mc\rightarrow\Mc'$ be an isometry. Then for every spectral filter $h$ and every $f\in\Lb^2(\Mc),$
\begin{equation*}
    T_h' V_\zeta(f) = V_\zeta T_h f \, .
\end{equation*}
\end{thm}
  
\begin{proof}
For $\lambda\in\Lambda,$ let $\pi_\lambda$ be the operator which projects a function $f\in \Lb^2(\Mc)$ onto the corresponding eigenspace $E_\lambda,$ and let $\pi_\lambda'$ be the analogous operator defined on $\Lb^2(\Mc').$ Since $\{\varphi_k\}_{\lambda_k=\lambda}$ forms an orthonormal basis for $E_\lambda$, we may write write $\pi_\lambda$ as integration against the kernel $K^{(\lambda)}(x,y)$ defined in \eqref{eqn: Klambda definition}, i.e.,
\begin{equation*}
    \pi_\lambda f(x) = \int_\Mc K^{(\lambda)}(x,y)f(y)dy \, .
\end{equation*}
Recalling from \eqref{eqn: splitupK} that 
\begin{equation*}
    K_h (x,y) = \sum_{\lambda\in\Lambda}H(\lambda)K^{(\lambda)}(x,y) \, ,
\end{equation*}
we see that
\begin{equation*}
    T_h f = \sum_{\lambda\in\Lambda} H(\lambda)\pi_\lambda f \, ,
\end{equation*}
and likewise,
\begin{equation*}
    T_h' f = \sum_{\lambda\in\Lambda} H(\lambda)\pi'_\lambda f \, .
\end{equation*}
Therefore, by the linearity of $V_\zeta,$ it suffices to show that
\begin{equation*}
    \pi'_\lambda V_\zeta f = V_\zeta \pi_\lambda f
\end{equation*} 
for all $f\in\Lb^2(\Mc)$ and all $\lambda\in\Lambda.$ Let $f\in \Lb^2(\Mc)$ and write 
\begin{equation*}
    f=f_1+f_2 \, ,
\end{equation*}
where $f_1\in E_\lambda, f_2\in E_\lambda^\perp.$ Since $\zeta$ is an isometry, we have $V_\zeta f_1\in E_\lambda'$ and $V_\zeta f_2\in \left(E_\lambda'\right)^\perp.$
Therefore, 
\begin{equation*}
    \pi_\lambda'V_\zeta f = \pi_\lambda' V_\zeta f_1 + \pi_\lambda' V_\zeta f_2 = V_\zeta f_1 = V_\zeta \pi_\lambda f
\end{equation*} 
as desired.
\end{proof}

\section{Proof of Proposition \ref{prop: lp frame}} \label{sec: pf of prop: lp frame}

\lpframe*

\begin{proof}
Analogously to Parseval's theorem, it follows from the Fourier inversion formula \eqref{eqn: fourier inversion} and the fact that $\{\varphi_k\}_{k\geq 0}$ is an orthonormal basis, that 
\begin{equation*}
    \|f\|_2^2 = \sum_{k\geq 0} |\widehat{f}(k)|^2 \, .
\end{equation*}
Similarly, it follows from \eqref{eqn: convolution on M} that
\begin{equation*} \label{eqn: int op norm in onb}
    \|  f \ast h_\gamma \|_2^2 = \sum_{k \geq 0} |\widehat{h_\gamma}(k)|^2 |\widehat{f}(k)|^2 \, .
\end{equation*}
Therefore, using the Littlewood Paley condition \eqref{eqn: lp type frame condition}, we see 
\begin{align*}
\| \mathcal{H} f \|_{2,2}^2 &=   \sum_{\gamma \in \Gamma} \|  f\ast h_\gamma \|_2^2 \\
&=  \sum_{\gamma \in \Gamma} \sum_{k \geq 0} |\widehat{h_\gamma}(k)|^2 |\widehat{f}(k)|^2 \\
&=  \sum_{k \geq 0} |\widehat{f}(k)|^2 \sum_{\gamma \in \Gamma} |\widehat{h_\gamma}(k)|^2  \\
&=  \sum_{k \geq 0} |\widehat{f}(k)|^2\\
&= \|f\|^2_2 \, .
\end{align*}

\end{proof}

\section{Proof of Proposition \ref{prop: wavelet isometry}}\label{sec: proof pf of prop: wavelet isometry}

\waveisom*

\begin{proof}
We will show that the frame  $\{\phi_J \, , \, \psi_j:j\leq J\}$ satisfies the Littlewood Paley condition \eqref{eqn: lp type frame condition}, i.e. that
\begin{equation*} \label{eqn: wavelet isometry proof 01}
    |\widehat{\phi}_J(k)|^2 + \sum_{j \leq J} |\widehat{\psi}_j (k)|^2 = 1, \quad \forall \, k\geq 0 \, .
\end{equation*} 
The result will then follow from Proposition \ref{prop: lp frame}. Recall that $\phi_J$ is defined by $\widehat{\phi}_J(k) = G\left(2^J\lambda_k\right)$ for some non-negative, non-increasing function $G$ such that $G(0) = 1$. Therefore, from (\ref{eqn: telescope}), we see that that 
\begin{equation*}
    |\widehat{\psi}_j (k)|^2 =  |\widehat{\phi}_{j-1} (k)|^2 - |\widehat{\phi}_{j} (k)|^2 = |G(2^{j-1}\lambda_k)|^2 - |G(2^j\lambda_k)|^2,
\end{equation*}
and so, 
\begin{align*}
 |\widehat{\phi}_J(k)|^2 + \sum_{j \leq J} |\widehat{\psi}_j (k)|^2 &= \left|G\left(2^J \lambda_k \right)\right|^2 + \sum_{j \leq J} \left[ \left|G\left(2^{j-1} \lambda_k \right)|^2 - |G\left(2^j \lambda_k \right)\right|^2 \right] \\
&= \lim_{j \rightarrow -\infty} |G \left(2^j \lambda_k \right)|^2 \\
&= |G(0)|^2 = 1, \quad \forall \, k \geq 0 \, .
\end{align*}
\end{proof}

\section{The Proof of Theorem \ref{thm: isometric invariance}} \label{sec: proof of isometry invariance}

\isoinv*

The proof of  Theorem \ref{thm: isometric invariance} relies on the following two lemmas.

\begin{lem} \label{lem: filter then diff stability}
There exists a constant $C (\Mc) > 0$ such that for every spectral filter $h$ and for every $\zeta \in \Diff (\Mc)$,
\begin{equation*}
    \| T_h f - V_{\zeta} T_h f \|_2 \leq C (\Mc) \left( \sum_{k \geq 0} \hh (k) \lambda_k^{d/2} \right) \| \zeta \|_{\infty} \| f \|_2 \, , \quad \forall \, f \in \Lb^2 (\Mc) \, .
\end{equation*}
Moreover, if $|\hh (k)| \leq e^{-2^J \lambda_k},$ then there exists a constant $C (\Mc) > 0$ such that for any $\zeta \in \Diff (\Mc)$, 
\begin{equation*}
    \| T_h f - V_{\zeta} T_h f \|_2 \leq  C (\Mc) 2^{-dJ} \| \zeta \|_{\infty} \| f \|_2 \, , \quad \forall \, f \in \Lb^2 (\Mc) \, .
\end{equation*}
\end{lem}

\begin{lem}\label{lem: UJmbound}
For any $f \in \Lb^2 (\Mc)$,
\begin{equation*} \|U_J^Lf\|_{2,2}\leq (L+1)^{1/2}\|f\|_2 \, .
\end{equation*}
\end{lem}

\begin{proof}[The Proof of Theorem \ref{thm: isometric invariance}]
Theorem \ref{thm: isometry equivariance} proves that spectral filter convolution operators commute with isometries. Since the absolute value operator does as well, it follows that  
$V_\zeta S_J^L=S_J^LV_\zeta,$ and therefore
\begin{equation*}
    \| S_J^L f - S_J^L  V_{\zeta}f \|_{2,2} = \| S_J^L f - V_{\zeta} S_J^L f \|_{2,2} \, .
\end{equation*}
Since $S_J^L  = A_J U_J^L $, we see that
\begin{align}
    \| S_J^L f - V_{\zeta} S_J^L f \|_{2,2} &= \| A_J U_J^L f - V_{\zeta} A_J U_J^L f \|_{2,2}
    \leq \| A_J - V_{\zeta} A_J \| \| U_J^L f \|_{2,2} \, . \label{eqn: S to U bound}
\end{align}
Since $A_J=T_{\phi_J}$ and $|\widehat{\phi_J}(k)| \leq  e^{-2^J \lambda_k}$, Lemma \ref{lem: filter then diff stability}  shows that 
\begin{equation*}
\| A_J - V_{\zeta} A_J \| \leq  C(\Mc) 2^{-Jd} \| \zeta \|_{\infty} \, .
\end{equation*}
Equation \eqref{eqn: isometric invariance for finite depth network} follows from Lemma \ref{lem: UJmbound}, and \eqref{eqn: isometric invariance for infinite depth network} follows by letting $L$ increase to infinity in \eqref{eqn: S to U bound}. Therefore, the proof is complete pending the proof of Lemmas \ref{lem: filter then diff stability} and \ref{lem: UJmbound}. 
\end{proof}

\begin{proof}[The Proof of Lemma \ref{lem: UJmbound}]
Let 
\begin{align*}
    \widetilde{U}_J^L f &\coloneqq\{ U_J^L [j_1, \ldots, j_L] f   :  \, j_i \leq J \enspace \forall \, 1 \leq i \leq L \} \\
    &=\{ |||f \ast \psi_{j_1}| \ast \psi_{j_2}| \ast \cdots \ast \psi_{j_{L}}| :  \, j_i \leq J \enspace \forall \, 1 \leq i \leq L \} \, .
\end{align*}
Then, by construction,
\begin{equation}\label{eqn: sum over path length}
    \|U_J^Lf\|_{2,2}^2= \sum_{\ell=0}^L \|\widetilde{U}_J^{\ell} f \|_{2,2}^2
\end{equation}
where we adopt the convention that $\widetilde{U}^0_J f=\{f\}.$ Since the wavelet transform and the absolute value operator are both non-expansive, it follows that $\widetilde{U}_J^1$ is non-expansive as well. Therefore, since $\widetilde{U}_J^{L}=\widetilde{U}_J^1 \widetilde{U}_J^{L-1},$ we see
\begin{equation*}
    \|\widetilde{U}_J^{L}f\|_{2,2} \leq \|\widetilde{U}_J^{L-1} f\|_{2,2} \leq \ldots \leq \|\widetilde{U}_J^{1} f\|_{2,2} \leq \|f\|_2 \, . 
\end{equation*}
Therefore, \eqref{eqn: sum over path length} implies
\begin{equation*}
    \|U_J^L\|_{2,2}^2= \sum_{\ell=0}^L \|\widetilde{U}_J^{\ell} \|_{2,2}^2\leq (L+1)\|f\|_2^2
\end{equation*}
as desired.
\end{proof}

In order to prove Lemma \ref{lem: filter then diff stability}, we will first prove the following lemma.

\begin{lem} \label{lem: kernel grad Linf bound}
For $\lambda\in\Lambda,$ let $K^{(\lambda)}$ be the kernel defined as in $\eqref{eqn: Klambda definition},$ and let $m(\lambda)$ be the multiplicity of $\lambda$. Then, there exists a constant $C(\Mc) > 0$ such that 
\begin{equation}\label{eqn: Klambda grad Linf bound} 
    \left\| \nabla K^{(\lambda)} \right\|_{\infty} \leq C(\Mc) m (\lambda) \lambda^{d/2} \, , \quad \forall \, \lambda\in\Lambda \, .
\end{equation}
As a consequence, if $K_h$ is a spectral kernel, then
\begin{equation}\label{eqn: kernel Linf grad bound}
    \| \nabla K_h \|_{\infty} \leq C(\Mc) \sum_{\lambda \in \Lambda} H (\lambda) m(\lambda) \lambda^{d/2} = C (\Mc) \sum_{k\geq 0} \hh (k) \lambda_k^{d/2} \, .
\end{equation}
Furthermore, if $\Mc$ is homogeneous, i.e., if for all $x,y\in\Mc,$ there exists an isometry mapping $x$ to $y,$ then
\begin{equation}\label{eqn: Klambda grad Linfbound Homogeneous}
    \left\| \nabla K^{(\lambda)} \right\|_{\infty} \leq C(\Mc) m (\lambda) \lambda^{(d+1)/4}
\end{equation}
and thus,
\begin{equation}\label{eqn: kernel Linf grad bound Homogeneous}
    \| \nabla K_h \|_{\infty} \leq C (\Mc)\sum_{\lambda\in\Lambda}H(\lambda)m(\lambda)\lambda^{(d+1)/4}= C (\Mc) \sum_{k \in \N} \hh (k) \lambda_k^{(d+1)/4} \, .
\end{equation}
\end{lem}

\begin{proof}
For any $k$ such that $\lambda_k=\lambda,$ it is a consequence of H\"ormander's local Weyl law (\cite{hormander1968}; see also \cite{shi:gradEigfcnManifold2010}) that
\begin{align*}
    \| \varphi_k \|_{\infty} &\leq C(\Mc) \lambda^{(d-1)/4} \, . \label{eqn: eigfcn Linf bound}
\end{align*}
Theorem 1 of \cite{shi:gradEigfcnManifold2010} shows that 
\begin{align*}
    \| \nabla \varphi_k \|_{\infty} &\leq C(\Mc) \sqrt{\lambda} \| \varphi_{k} \|_{\infty} \, . 
\end{align*}
Therefore,
\begin{align*}
    \left|\nabla K^{(\lambda)} (x,y)\right|^2 &= \left| \sum_{k : \lambda_k = \lambda} \nabla \varphi_k (x) \overline{\varphi}_k (y) \right|^2 \\
    &\leq \left( \sum_{k : \lambda_k = \lambda} |\nabla \varphi_k (x)|^2 \right) \left( \sum_{k : \lambda_k = \lambda} |\varphi_k (y)|^2 \right) \\
    &\leq C(\Mc) m (\lambda) \lambda^{(d-1)/2} \sum_{k : \lambda_k = \lambda} |\nabla \varphi_k (x)|^2 \\
    &\leq C (\Mc) m(\lambda) \lambda^{(d+1)/2} \sum_{k : \lambda_k = \lambda} \| \varphi_k \|_{\infty}^2 \\
    &\leq C(\Mc) m(\lambda)^2 \lambda^d \, .
\end{align*}
This implies \eqref{eqn: Klambda grad Linf bound}. Now, if we assume that $\Mc$ is homogeneous, then Theorem 3.2 of \cite{evarist1975addition} shows that 
\begin{equation*}
    \sum_{k : \lambda_k = \lambda} |\varphi_k (y)|^2= C(\Mc)m(\lambda) \, .
\end{equation*}
Substituting this into the above string of inequalities yields \eqref{eqn: Klambda grad Linfbound Homogeneous}. Equations \eqref{eqn: kernel Linf grad bound} and \eqref{eqn: kernel Linf grad bound Homogeneous} follow by recalling from \eqref{eqn: splitupK} that 
\begin{equation*}
    K_h (x,y) =\sum_{\lambda\in\Lambda} H(\lambda)K^{(\lambda)}(x,y) \, ,
\end{equation*}
and applying the triangle inequality.
\end{proof}

Now we may prove Lemma \ref{lem: filter then diff stability}.

\begin{proof}[The Proof of Lemma \ref{lem: filter then diff stability}]
Let $K_h$ be the kernel of $T_h$. Then  by the Cauchy-Schwartz inequality and the fact that  $V_\zeta f(x) = f\left(\zeta^{-1}(x)\right),$ 
\begin{align*}
    | T_h f(x) - V_{\zeta} T_h f(x) | &= \left| \int_{\Mc} \left[K_h(x,y) - K_h\left(\zeta^{-1}(x),y\right)\right] f(y) \, dy \right| \\
    &\leq \| f \|_2 \left( \int_{\Mc} \left|K_h(x,y) - K_h\left(\zeta^{-1}(x),y\right)\right|^2 \, dy \right)^{1/2} \\
    &\leq \| f \|_2 \| \nabla K_h \|_{\infty}\left( \int_{\Mc}  \left|r\left(x, \zeta^{-1} (x)\right)\right|^2 \, dy \right)^{1/2} \\
    &\leq \| f \|_2 \sqrt{\vol (\Mc)} \| \nabla K_h \|_{\infty} \| \zeta \|_{\infty} \, .
\end{align*}
It follows that  
\begin{equation} \label{eqn: genlemma6}
    \| T_h f- V_{\zeta} T_h f\|_2 \leq \vol (\Mc) \| \nabla K_h \|_{\infty} \| \zeta \|_{\infty} \, .
\end{equation}
Lemma \ref{lem: kernel grad Linf bound} shows 
\begin{equation*}
    \| \nabla K_h \|_{\infty} \leq C(\Mc) \sum_{k\geq 0} \widehat{h} (k) \lambda_k^{d/2} \, ,
\end{equation*}
and therefore
\begin{equation*}
    \| T_h f- V_{\zeta} T_h f\|_2 \leq C(\Mc) \left( \sum_{k\geq 0} \widehat{h} (k) \lambda_k^{d/2} \right) \| \zeta \|_{\infty} \, .
\end{equation*}

Now suppose that $G(\lambda)\leq e^{-\lambda}$. Theorem 2.4 of \cite{berard:embedManifoldHeatKer1994} proves that for any $x \in \Mc$, $\alpha \geq 0$, and $t > 0$,
\begin{equation}\label{eqn: sum over eigs exponential}
    \sum_{k \geq 1} \lambda_k^{\alpha} e^{-t \lambda_k} |\varphi_k(x)|^2 \leq C (\Mc) (\alpha + 1) t^{-(d + 2\alpha)/2} \, .
\end{equation}
Integrating both sides over $\Mc$ yields:
\begin{equation} \label{eqn: weighted heat kernel eig sum}
    \sum_{k \geq 1} \lambda_k^{\alpha} e^{-t \lambda_k} \leq C(\Mc) (\alpha + 1) t^{-(d + 2\alpha)/2} \, .
\end{equation}
Using the assumption that that $|\hh (k)| \leq  e^{-2^J \lambda_k},$  (\ref{eqn: genlemma6})  and \eqref{eqn: weighted heat kernel eig sum} with $\alpha = d/2$ and $t=2^J,$ we see
\begin{align*}
\| T_hf- V_{\zeta} T_hf \|_2 &\leq C(\Mc) \left( \sum_{k \geq 1} \lambda_k^{d/2} e^{-2^J \lambda_k} \right) \| \zeta \|_{\infty} \\
&\leq C(\Mc) 2^{-dJ} \| \zeta \|_{\infty} \, .
\end{align*}

\end{proof}

\section{The Proof of Theorem \ref{prop: littleelltwo}}
\label{sec: Proof of Proposition littleelltwo}
\littleelltwo*

\begin{proof}
Let $p = \emptyset$ or $p = (j_1, \ldots, j_m)$ denote a scattering path. 
%, and let 
% \begin{equation} \label{eqn: all scattering paths}
%     \Pc := \{ \emptyset \} \cup \{ (j_1, \ldots, j_m) : m \geq 1 \, , \, j_i \in \Z \enspace \forall \, 1 \leq i \leq m \}
% \end{equation}
% denote the set of all scattering paths. Denote by $\Pc_J$ the set of all paths with scales no larger than $J$,
% \begin{equation*}
%     \Pc_J := \{ \emptyset \} \cup \{ (j_1, \ldots, j_m) : m \geq 1 \, , \, j_i \leq J \enspace \forall \, 1 \leq i \leq m \} \, .
% \end{equation*}
Using \eqref{eqn: J infinity}, 
implies
\begin{equation*}
    \lim_{J\rightarrow \infty}\left\|\frac{|S_J[p] f|}{
    \vol(\Mc)^{-1/2}
    }-\overline{S}[p]f\right\|_2=0,
\end{equation*}
and therefore
\begin{equation*}\lim_{J\rightarrow \infty}\left\|\frac{|S_J[p] f|}{\vol(\Mc)^{-1/2}}\right\|_2={\vol(\Mc)^{1/2}}\overline{S}[p]f.
\end{equation*}

Thus, using Fatou's lemma, we have
\begin{align*}\|\overline{S}f\|_2^2&=\sum_p |\overline{S}[p]f|^2\\
&=\frac{1}{\vol(\Mc)}\sum_p\lim_{J\rightarrow\infty}\left\|\frac{|S_J[p]f|}{\vol(\Mc)^{-1/2}}\right\|_2^2\\
&\leq \liminf_{J\rightarrow\infty}\sum_p\left\||S_J[p]f|\right\|_2^2\\
%&\leq \liminf_{J\rightarrow\infty}\sum_p\left\|S_J[p]f\right\|_2^2\\
%&\leq \liminf_{J\rightarrow\infty}\sum_p\left\|S_J[p]f\right\|_2^2\\
&\leq \|f\|_2,
\end{align*}
where in the last line we applied Proposition \ref{prop: nonexpansive}.

\end{proof}

\section{The Proof of Proposition \ref{prop: nonwindowedinvdiff}}
\label{Sec: The proof of Prop: nonwindowedinvdiff}

\nonwindowinvariancedifferent*

\begin{proof}
Letting $p = \emptyset$ or $p = (j_1, \ldots, j_m)$ denote a scattering path, we see that we need to prove $\Sbar [p] f = \Sbar' V_{\zeta} f[p]$ for all $p \in \Pc$. If $p = \emptyset$ then $\Sbar f (\emptyset) = \| f \|_{\Lb^1 (\Mc)} = \| V_{\zeta} f \|_{\Lb^1 (\Mc')} = \Sbar' V_{\zeta} f (\emptyset)$ since $\zeta$ is an isometry. Theorem \ref{thm: equivariance, different mfolds}, stated in Appendix \ref{sec: pf of isometry equivariance}, proves that $T_h' V_{\zeta} f = V_{\zeta} T_h f$ for any spectral filter $h \in \Lb^2 (\Mc)$, where $T_h'$ is the analogue of $T_h$ on $\mathbf{L}^2(\Mc')$ (defined precisely in Appendix \ref{sec: pf of isometry equivariance}). Since the modulus operator also commutes with isometries, it follows that $U'[p] V_{\zeta} f = V_{\zeta} U[p] f$ for any $p \in \Pc.$ Thus, since $\zeta$ is an isometry,
\begin{equation*}
    \Sbar' V_{\zeta} f (p) = \| U'[p] V_{\zeta} f \|_{\Lb^1 (\Mc')} = \| V_{\zeta} U[p] f \|_{\Lb^1 (\Mc')} = \| U [p] f \|_{\Lb^1 (\Mc)} = \Sbar f (p) \, .
\end{equation*}
\end{proof}

\section{The Proof of Theorem \ref{thm: isoinvariancediff}}
\label{sec: proof for isometrydiff}

\isoinvariancediff*

\begin{proof}
As in the proof of Theorem \ref{thm: isometric invariance}, we observe that since spectral filter convolution operators and the absolute value operator both commute with isometries, it follows that $\left(S_J^L\right)' V_{\zeta_1}=V_{\zeta_1} S_J^L.$ Therefore
\begin{equation*}
    \| S_J^L  f - V_{\zeta_2^{-1}}\left(S_J^L\right)'  V_{\zeta_1} f\|_{2,2} = 
    \|  S_J^L  f - V_{\zeta_2^{-1}} V_{\zeta_1}S_J^L f\|_{2,2}= \|S_J^Lf-V_{\zeta_2^{-1}\circ\zeta_1}S_J^L f\|_{2,2} \, .
\end{equation*}
Equation \eqref{eqn: isometric invariance different finite} now follows by applying \eqref{eqn: isometric invariance for finite depth network}. The proof of \eqref{eqn: isometric invariance different Infinite} is similar and follows by applying \eqref{eqn: isometric invariance for infinite depth network}. 
\end{proof}

\section{The Proof of Theorems \ref{thm: smudgestablityBL} and \ref{thm: smudgestablityBLnowindow}}
\label{sec: proof of blstability}

\smudgestabilityBL*

\smudgestabilityBLnowindow*
In order to prove Theorem \ref{thm: smudgestablityBL}, we will need the following lemma.

\begin{lem} \label{lem: supboundBL}
If $f\in\Lb^2(\Mc)$ is $\lambda$-bandlimited, i.e., $\langle f,\varphi_k\rangle=0$ whenever $\lambda_k>\lambda,$  then there exists a constant $C(\Mc)<\infty$ such that
\begin{equation*}
    \|f-V_\zeta f\|_2 \leq C(\Mc)\lambda^{d}\|\zeta\|_\infty\|f\|_2 
\end{equation*}
for all $\zeta \in \Diff (\Mc)$.
\end{lem}

\begin{proof}
As in the proof of Theorem \ref{thm: equivariance, different mfolds}, let $\Lambda$ denote the set of unique eigenvalues of $-\Delta,$ and let $\pi_\lambda$ be the operator that projects a function $f\in\Lb^2(\Mc)$ onto the eigenspace $E_\lambda.$ Let 
\begin{equation*}
    P_\lambda \coloneqq \sum_{\tilde{\lambda}\leq \lambda}\pi_{\tilde{\lambda}} \, ,
\end{equation*}
be the operator which projects a function $f\in\Lb^2(\Mc)$ onto all eigenspaces with eigenvalues less than or equal to $\lambda.$ Note that $P_\lambda$ can be written as integration against the kernel 
\begin{equation*}
    K(x,y)=\sum_{\lambda_k\leq \lambda} \varphi_k(x)\overline\varphi_k(y)=\sum_{\tilde{\lambda}\leq\lambda}K^{(\lambda)}(x,y) \, ,
\end{equation*}
where $K^{(\lambda)}$ is defined as in \eqref{eqn: Klambda definition}. If $f$ is any $\lambda$-bandlimited function in $\Lb^2(\Mc)$, then $P_\lambda f = f,$ and so similarly to the proof of Lemma \ref{lem: filter then diff stability}, we see that
\begin{align*}
    |f(x)-V_\zeta f(x)|&=|P_\lambda f(x) -V_\zeta P_\lambda f (x)|\\
    &=\left|\int_\Mc K(x,y) f(y) dy- \int_\Mc K\left(\zeta^{-1}(x),y\right)f(y)dy\right|\\
    &\leq \|f\|_2\left(\int_\Mc \left|K(x,y) - K\left(\zeta^{-1}(x),y\right)\right|^2 dy\right)^{1/2}\\
    &\leq \|f\|_2 \|\zeta\|_\infty \sqrt{\vol(\Mc)} \|\nabla K\|_\infty \, ,
\end{align*}
which implies
\begin{equation*}
    \|f-V_\zeta f\|_2 \leq \vol(\Mc)\|\nabla K\|_\infty\|\zeta\|_\infty\|f\|_2 \, .
\end{equation*}
Lemma \ref{lem: kernel grad Linf bound} shows that for all $\tilde{\lambda}$
\begin{equation*}
    \left\| \nabla K^{(\tilde{\lambda})} \right\|_{\infty} \leq C(\Mc) m (\tilde{\lambda}) \tilde{\lambda}^{d/2} \, .
\end{equation*}
Therefore, 
\begin{equation*}
    \|\nabla K\|_\infty \leq C(\Mc) \sum_{\lambda_k\leq \lambda} \left(\lambda_k\right)^{d/2} \leq C(\Mc) N (\lambda) \lambda^{d/2} \, ,
\end{equation*}
where $N (\lambda)$ is the number of eigenvalues less than or equal to $\lambda.$ Weyl's law (see for example \cite{Ivrii2016}) implies that 
\begin{equation*}
    N (\lambda) \leq C(\Mc)\lambda^{d/2} \, ,
\end{equation*}
and so
\begin{equation*}
    \|\nabla K\|_\infty  \leq C(\Mc)  \lambda^d \, .
\end{equation*}
\end{proof}

\begin{proof}[The Proof of Theorem \ref{thm: smudgestablityBL}]
Let $\zeta=\zeta_2\circ\zeta_1$ be a factorization of $\zeta$ such that $\zeta_1$ is an isometry  and $\zeta_2$ is a diffeomorphism. Then since $V_\zeta f = f\circ \zeta^{-1},$ we see that $V_\zeta = V_{\zeta_2}V_{\zeta_1}.$ Therefore, for all $\lambda$-bandlimited functions $f$
\begin{equation*}
    \|S_J^Lf-S_J^LV_\zeta f\|_{2,2} \leq \|S_J^Lf - S_J^LV_{\zeta_1}f\|_{2,2} + \|S_J^LV_{\zeta_1}f - S_J^LV_{\zeta_2}V_{\zeta_1}f\|_{2,2} \, . 
\end{equation*}
By \eqref{eqn: isometric invariance for finite depth network}, we have that 
\begin{equation*}
    \| S_J^L  f - S_J^L  V_{\zeta_1} f \|_{2,2} \leq C(\Mc) 2^{-Jd} \| \zeta_1 \|_{\infty}(L+1)^{1/2} \| f \|_2 \, ,  
\end{equation*}
and by Proposition \ref{prop: nonexpansive} and Lemma \ref{lem: supboundBL} we see
\begin{equation*}\|S_J^LV_{\zeta_1}f - S_J^LV_{\zeta_2}V_{\zeta_1}f\|_{2,2}\leq
    \|V_{\zeta_1}f-V_{\zeta_2} V_{\zeta_1}f\|_2 \leq C(\Mc)\lambda^{d}\|\zeta_2\|_\infty\|V_{\zeta_1}f\|_2 \, .
\end{equation*}
Since $\zeta_1$ is an isometry, we observe that $\|V_{\zeta_1}f\|_2=\|f\|_2.$ Combining this with the two inequalities above completes the proof of \eqref{eqn: smudgestabilityfinite}. The proof of \eqref{eqn: smudgestabilityinfinity} is similar, but uses \eqref{eqn: isometric invariance for infinite depth network} instead of \eqref{eqn: isometric invariance for finite depth network}.
\end{proof}

\begin{proof}[The Proof of Theorem \ref{thm: smudgestablityBLnowindow}]
Repeating the proof of Proposition \ref{prop: nonwindowedinvdiff}
we see that 
\begin{equation*} 
    \|\overline{S} f - \overline{S} V_\zeta f \|_{2,2}\leq \lim_{J\rightarrow\infty} \|S_J f - S_J V_\zeta f \|_{2,2}.
\end{equation*}
Therefore, the result follows from Theorem \ref{thm: smudgestablityBL} by taking $J \rightarrow \infty$ on the right hand side of \eqref{eqn: smudgestabilityinfinity}.
\end{proof}

\section{The Proof of Theorem \ref{twoptho}} 
\label{sec: commutator proof}

\twopt*

In order to prove Theorem \ref{twoptho}, we will need the an auxiliary result, which provides a commutator estimate for operators with radial kernels. We will say that a kernel operator 
\begin{equation*}
    Tf(x) = \int_\Mc K(x,y)f(y) \, dy
\end{equation*}
is radial if 
\begin{equation*}
    K (x,y)=\kappa(r(x,y))
\end{equation*}
for some $\kappa:[0,\infty)\rightarrow \mathbb{C}.$ The following theorem establishes a commutator estimate for operators with radial kernels.
 
\begin{thm} \label{thm: dist kernel commutator}
Let $T$ be a kernel integral operator with a radial kernel  $K(x,y) = \kappa (r(x,y))$ for some $\kappa \in \Cb^1 (\R)$. Then there exists constants $C_1 (\Mc, K)$ and $C_2 (\Mc, K)$ such that
\begin{equation*}
    \| [T,V_\zeta] f \|_2 \leq \|f\|_2 \left[ C_1 (\Mc, K) A_1(\zeta) + C_2 (\Mc, K) A_2(\zeta) \right] \, .
\end{equation*}
Here $A_1(\zeta)$ and $A_2(\zeta)$ are defined as in (\ref{a1}) and (\ref{a2}) respectively,
\begin{equation*}
    C_1 (\Mc, K) = \| \nabla K \|_{\infty} \diam (\Mc) \vol (\Mc) \, ,
\end{equation*}
and
\begin{equation*}
    C_2 (\Mc, K) = \| K \|_{\Lb^2 (\Mc \times \Mc)} \, .
\end{equation*}
\end{thm}

\begin{proof}
We first compute
\begin{align*}
    |[T,V_\zeta]f(x)| &= \left|\int_{\Mc} K(x,y)f(\zeta^{-1}(y)) \, dy - \int_{\Mc} K(\zeta^{-1}(x),y)f(y) \, dy\right| \\
    &=\left|\int_{\Mc} K(x,\zeta(y))f(y)|\det[D\zeta (y)]| \, dy - \int_{\Mc} K(\zeta^{-1}(x),y)f(y) \, dy\right| \\
    &=\left| \int_{\Mc} f(y) \left[ K (x,\zeta(y))|\det[D\zeta (y)]|- K(\zeta^{-1}(x),y) \right] \, dy \right| \\
    &\leq \left| \int_{\Mc} f(y) \left[ K(x,\zeta(y))[|\det[D\zeta (y)]|-1 \right] \, dy \right| + \left| \int_{\Mc} f(y) \left[ K(x,\zeta(y))- K (\zeta^{-1}(x),y) \right] \, dy \right| \\
    &\leq \||\det[D\zeta(y)]|-1\|_\infty \left| \int_{\Mc} f(y) K (x,\zeta(y)) \, dy \right| + \left| \int_{\Mc} f(y) \left[ K(x,\zeta(y))-K(\zeta^{-1}(x),y) \right] \, dy \right|.
\end{align*}
Therefore, by the Cauchy-Schwartz inequality,
\begin{align*}
\|[T,V_\zeta]f\|_2 \leq \|f\|_2 \Bigg[ &\||\det[D\zeta(y) ]|-1\|_\infty \bigg(\int_{\Mc} \int_{\Mc} \left|K(x,\zeta(y))\right|^2 \, dy \, dx \bigg)^{\frac{1}{2}} \\
&+ \bigg( \int_{\Mc} \int_{\Mc} \left| K (x,\zeta(y))- K (\zeta^{-1}(x),y)\right|^2 \, dy \, dx \bigg)^{\frac{1}{2}} \Bigg] \, .
\end{align*}

We may bound the first integral by observing
\begin{equation*}
    \int_{\Mc} \int_{\Mc} \left| K (x,\zeta(y)) \right|^2 \, dy \, dx \leq \|\det[D\zeta^{-1}(y)]\|^2_\infty \int_{\Mc} \int_{\Mc} \left| K(x,y)\right|^2 \, dy \, dx \, .
\end{equation*}
To bound the second integral observe, that by the mean value theorem and the assumption that $K$ is radial, we have
\begin{align*}
\int_{\Mc} \int_{\Mc} |K (x,\zeta(y)) &- K (\zeta^{-1}(x),y)|^2 \, dy \, dx \\
    &= \int_{\Mc} \int_{\Mc} \left|\kappa (r(x,\zeta(y)))- \kappa(r(\zeta^{-1}(x),y))\right|^2 \, dy \, dx \\
    &\leq \|\kappa'\|_\infty^2 \int_{\Mc} \int_{\Mc} \left|r(x,\zeta(y))- r(\zeta^{-1}(x),y)\right|^2 \, dy \, dx \\
    &\leq \left[ \|\kappa'\|_\infty A_1(\zeta) \right]^2 \int_{\Mc} \int_{\Mc} \left| r(x,\zeta(y)) \right|^2 \, dy \, dx \\
    &\leq \left[ \|\kappa'\|_\infty A_1(\zeta) \diam(\Mc) \vol(\Mc) \right]^2.
\end{align*}
Lastly, since $K(x,y)=\kappa(r(x,y)),$ we see that 
\begin{equation*}
\|\nabla K\|_\infty = \|\kappa'\|_\infty \, ,
\end{equation*}
which completes the proof.
\end{proof}

\begin{proof}[The Proof of Theorem \ref{twoptho}]
We write $T = T_h$ and $K = K_h$. If $\Mc$ is two-point homogeneous  
and $r(x, y) = r(x', y'),$ then by the definition of two-point homogeneity there exists an isometry $\widetilde{\zeta}$ mapping $x \mapsto x'$ and $y \mapsto y'$. Therefore, we may use the proof of Theorem \ref{thm: isometry equivariance} to see that $K(x',y') = K(x,y)$. It follows that $K (x,y)$ is radial and so we may write $K (x,y) = \kappa (r(x,y))$ for some $\kappa\in\Cb^1$. 

Applying Theorem \ref{thm: dist kernel commutator}, we see that
\begin{equation*}
    \| [T, V_{\zeta}] \| \leq C (\Mc) \left[ \| \nabla K \|_{\infty} A_1 (\zeta) + \| K \|_{\Lb^2 (\Mc \times \Mc)} A_2(\zeta)\right].
\end{equation*}
Lemma \ref{lem: kernel grad Linf bound} implies that
\begin{equation*}
    \| \nabla K \|_{\infty} \leq C (\Mc) \sum_{k \in \N} H (\lambda_k) \lambda_k^{(d+1)/4} \, ,
\end{equation*}
and since $\{\varphi_k\}_{k=0}^\infty$ forms an orthonormal basis for $\Lb^2(\Mc),$ it can be checked that  
\begin{equation*}
    \|K\|_{\Lb^2(\Mc\times\Mc)} = \left(\sum_{k=0}^\infty |H(\lambda_k)|^2\right)^{1/2} \, .
\end{equation*}
Therefore, the proof is complete since 
\begin{equation*}
    A(\zeta)=\max\{A_1(\zeta),A_2(\zeta)\} 
\end{equation*}
and
\begin{equation*}
     B (h) = \max \left\{ \sum_{k \in \N} H (\lambda_k) \lambda_k^{(d+1)/4}, \left( \sum_{k \in \N} H (\lambda_k)^2 \right)^{\frac{1}{2}} \right\} \, .
\end{equation*}
\end{proof}

\section{The Proof of Corollary \ref{cor: single wavelet stabilty}}
\label{sec: The Proof of Corollary  single wavelet stabilty}

\singlewavelet*

\begin{proof}
By the definition of $\hpsi_j$ and the assumption that $G(\lambda)\leq e^{-\lambda},$ 
we have that $\hpsi_j(0)=0$ and
\begin{equation*}
    |\hpsi_j(k)|\leq |\hphi_{j-1}(k)|\leq e^{-2^{j-1}\lambda_k} 
\end{equation*} 
for $k\geq 1.$ Therefore, by Theorem \ref{twoptho},
\begin{align*}
\| [\Psi_j, V_{\zeta}]\| &\leq C (\Mc) A (\zeta) B (\psi_j)
\end{align*}
where,
\begin{equation*}
B (\psi_j) = \max \left\{ \sum_{k \geq 1} e^{-2^{j-1}\lambda_k} \lambda_k^{(d+1)/4}, \left( \sum_{k\geq1} e^{-2^{j}\lambda_k} \right)^{\frac{1}{2}} \right\}.
\end{equation*}
Equation \eqref{eqn: sum over eigs exponential} implies that 
\begin{equation*}
    \sum_{k\geq1} e^{-2^{j-1}\lambda_k} \lambda_k^{(d+1)/4}\leq C(\Mc) 2^{-(d+1/2)(j-1)}
\end{equation*}
and that 
\begin{equation*}
    \sum_{k\geq1} e^{-2^{j}\lambda_k} \leq C(\Mc) 2^{-dj/2}
\end{equation*}
thus completing the proof.
\end{proof}

\section{Additional details of classification experiments}
\label{sec: numerical details}

\subsection{Spherical MNIST}

For all spherical MNIST experiments, including those reported in Section \ref{sec: mnist}, we used the following procedure. Since the digits six and nine are impossible to distinguish in spherical MNIST, we removed the digit six from the dataset. The mesh on the sphere consisted of 642 vertices, and to construct the wavelets on the sphere, all 642 eigenvalues and eigenfunctions of the approximate Laplace-Beltrami operator were used. For the range of scales we chose $-8 \leq j \leq \min (0,J)$. Training and testing were conducted using the standard MNIST training set of 60,000 digits and the standard testing set of 10,000 digits, projected onto the sphere. The training set was randomly divided into five folds, of which four were used to train the RBF kernel SVM, taking as input the relevant geometric scattering representation of each spherically projected digit, and one was used to validate the hyper-parameters of the RBF kernel SVM (see Appendix \ref{sec: rbf kernel svm}). 

On both the non-rotated and randomly rotated spherical MNIST datasets, we calculated the geometric scattering coefficients $S_J^L f$ and  downsampled the resulting scattering coefficient functions (e.g., $f \ast \phi_J (x)$ and $|f \ast \psi_j| \ast \phi_J (x)$). For $J \rightarrow \infty$ we selected one coefficient since they are all the same (recall \eqref{eqn: J infinity}). With $J= 0$, we selected 4 coefficients per function; with $J = -1$, we selected 16 coefficients; with $J = -2$, we selected 64 coefficients. The selected coefficients were determined by finding 4, 16, and 64 nearly equidistant points $x$ on the sphere. Classification results on the test set for a fixed network depth of $L = 2$ and for the different values of $J$ are reported in Table \ref{table: J/NR/R scattering mnist} below. Figure \ref{table: scattering order J inf} reports classification results for $J \rightarrow \infty$ and for $0 \leq L \leq 3$. 

\begin{table*}[!htb]
\caption{Geometric wavelet scattering classification results with with network depth $L=2$ for different $J$ on non-rotated and rotated spherical MNIST}
\centering
\begin{tabular}{|c|c|c|}
\hline
$J$ & NR & R \\
\hline
$J \rightarrow \infty$ & $0.91$ & $0.91$\\
\hline
$J=0$ & $0.94$ & $0.94$\\
\hline
$J=-1$ & $0.95$ & $0.95$\\
\hline
$J=-2$ & $0.95$ & $0.95$\\
\hline
\end{tabular}
\label{table: J/NR/R scattering mnist}
\end{table*}

\subsection{FAUST}
\label{app: faust}

The FAUST dataset \cite{Bogo:CVPR:2014} consists of $100$  manifolds corresponding to ten distinct people in ten distinct poses. Each manifold is approximated by a mesh with $6890$ vertices. We used the $512$ smallest eigenvalues and corresponding eigenfunctions to construct the geometric wavelets. During cross validation, in addition to cross validating the SVM parameters (see Section \ref{sec: rbf kernel svm} below), we also cross validated the depth of the scattering network for $0 \leq L \leq 2$. For the classification test, we performed $5$ fold cross validation with a training/validation/test split of 70\%/10\%/20\% for both pose classification and person classification. The range of $j$ is chosen as $-11 \leq j \leq 0$.

We report the frequency of each network depth $L$ selected during the hyperparameter cross validation stage. Since there are five test folds and eight validation folds, the depth is selected $40$ times per task. For pose classification, $L=0$ was selected $19$ times, $L=1$ was selected $11$ times, and $L=2$ was selected $10$ times. For person classification, $L=0$ was selected $5$ times, $L=1$ was selected $29$ times, and $L=2$ was selected $6$ times. The results indicate the importance of avoiding overfitting with needlessly deep scattering networks, while at the same time highlighting the task dependent nature of the network depth (compare as well to the MNIST results reported above and in the main text). 

\subsection{Parameters for RBF kernel SVM}
\label{sec: rbf kernel svm}

We used RBF kernel SVM for all classification tasks and cross validated the hyperparameters. In the two FAUST classification tasks of Section \ref{sec: faust}, for the kernel width $\gamma$, we chose from $\{ 0.001,0.005,0.01,0.02,0.04 \}$, while for the penalty $C$ we chose from $\{ 50,100,250,400,500 \}$. For the spherical MNIST classification task of Section \ref{sec: mnist}, for $\gamma$ we chose from $\{ 0.00001,0.0001,0.001 \}$ and for $C$ we chose from $\{ 25,100,250,500 \}$. %% <-- these are the appendices; comment out in two-file mode.
 
\end{document}